\documentclass[twoside]{article}

\usepackage[accepted]{aistats2022}
\usepackage{color}
\definecolor{mycitecolor}{rgb}{0,0.08,0.45}

\usepackage[round]{natbib}

\usepackage{url}
\usepackage{booktabs}
\usepackage{amsfonts}
\usepackage{nicefrac}
\usepackage{microtype}
\usepackage{multirow}

\usepackage{wrapfig}
\usepackage{graphicx}
\usepackage{comment}
\usepackage{amsmath,amssymb}

\usepackage{booktabs}
\usepackage{amsthm}
\usepackage{algorithm}
\usepackage{algorithmic}

\usepackage{symbols}
\usepackage{multirow}
\usepackage{pmboxdraw}
\usepackage{mathtools}
\usepackage[table,xcdraw,dvipsnames]{xcolor}
\usepackage{thm-restate}
\renewcommand{\cite}[1]{\citep{#1}}

\usepackage[colorlinks=true, citecolor=mycitecolor]{hyperref}

\usepackage{amsmath,amsfonts,bm}

\def\1{\bm{1}}

\def\eps{{\epsilon}}

\def\rva{{\mathbf{a}}}

\def\rvg{{\mathbf{g}}}

\def\rvx{{\mathbf{x}}}
\def\rvy{{\mathbf{y}}}

\def\vb{{\bm{b}}}

\def\mA{{\bm{A}}}

\def\mG{{\bm{G}}}

\def\mI{{\bm{I}}}

\def\mW{{\bm{W}}}
\def\mX{{\bm{X}}}
\def\mY{{\bm{Y}}}
\def\mZ{{\bm{Z}}}

\DeclareMathAlphabet{\mathsfit}{\encodingdefault}{\sfdefault}{m}{sl}
\SetMathAlphabet{\mathsfit}{bold}{\encodingdefault}{\sfdefault}{bx}{n}

\def\gC{{\mathcal{C}}}
\def\gD{{\mathcal{D}}}

\def\gG{{\mathcal{G}}}
\def\gH{{\mathcal{H}}}

\def\gP{{\mathcal{P}}}

\def\gS{{\mathcal{S}}}
\def\gT{{\mathcal{T}}}

\def\gV{{\mathcal{V}}}

\DeclareMathOperator*{\argmin}{arg\,min}

\usepackage[framemethod=TikZ]{mdframed}
\mdfdefinestyle{MyFrame}{
    linecolor=Black,
    outerlinewidth=0.3pt,
    roundcorner=3pt,
    innertopmargin=0.3\baselineskip,
    innerbottommargin=.3\baselineskip,
    innermargin = -0.3cm,
  	outermargin = 0cm,
    innerrightmargin=4.5pt,
    innerleftmargin=4.5pt,
    backgroundcolor=Gray!0!white}

\mdfdefinestyle{MyFrame2}{
    linecolor=White,
    outerlinewidth=.5pt,
    roundcorner=5pt,
    innertopmargin=\baselineskip,
    innerbottommargin=\baselineskip,
    innerrightmargin=5pt,
    innerleftmargin=5pt,
    backgroundcolor=Gray!10!white}

\newcommand {\MSE}[1]{\text{MSE}(#1)}
\newcommand {\hMSE}[1]{\widehat{\text{MSE}}(#1)}

\newcommand {\Rank}[1]{\text{rk}(#1)}

\usepackage[multiple]{footmisc}
\usepackage{ctable}
\usepackage{pifont}
\newcommand{\cmark}{\ding{51}}
\newcommand{\xmark}{\ding{55}}

\usepackage[utf8]{inputenc}
\usepackage{pmboxdraw}

\begin{document}

\twocolumn[

\aistatstitle{Equivariance Discovery by Learned Parameter-Sharing}
\aistatsauthor{Raymond A. Yeh\textsuperscript{$\dagger$}\;\;\;\;Yuan-Ting Hu\;\;\;\; Mark Hasegawa-Johnson\;\;\;\;  Alexander G. Schwing}

\aistatsaddress{Toyota Technological Institute at Chicago\textsuperscript{$\dagger$}\;\;\;\;\;\;\;University of Illinois at Urbana-Champaign} ]

\begin{abstract}

Designing equivariance as an inductive bias into deep-nets has been a prominent approach to build effective models, \eg, a convolutional neural network incorporates translation equivariance. 
However, incorporating these inductive biases requires knowledge about the equivariance properties of the data, which may not be available, \eg, when encountering a new domain. To address this, we study how to \textit{discover interpretable equivariances} from data. 
Specifically, we formulate this discovery process as an optimization problem over a model's parameter-sharing schemes. We  propose to use the partition distance to empirically quantify the accuracy of the recovered equivariance. Also, we theoretically analyze the method for Gaussian data and provide a bound on the 
mean squared  gap between the studied discovery scheme and the oracle scheme. 
Empirically, we show that the approach recovers known equivariances,
such as permutations and shifts,
on sum of numbers and spatially-invariant data.
\end{abstract}

\section{INTRODUCTION} %
Encoding equivariance and invariance into deep-nets has been an effective method to improve the data-efficiency of machine learning models. For example, convolutional neural nets (CNNs) or recurrent neural nets (RNNs) encode shift-equivariant properties either in the spatial or temporal domain~\cite{lecun1999object, hochreiter1997long}. More recently, equivariance has also been studied in other domains,~\eg, over sets, graphs, or other geometric structures~\citep{zaheer2017deep, bronstein2017geometric}. 

While encoding equivariance has been remarkably successful, encoding requires a-priori knowledge about the desirable equivariance properties to be built into a model. Such knowledge of equivariance requires domain expertise which may not be available. To tackle this concern, we study \textit{discovery of interpretable equivariance} from data rather than manually imposing it.

For this, we consider discovery of equivariance over a family of discrete group actions. This family of equivariances can be built into deep-nets via parameter-sharing~\cite{ravanbakhsh2017equivariance}. In other words, if we can learn how to share parameters, then we can discover equivariance. To achieve this goal, we identify a parametric representation of the parameter-sharing scheme. This permits to cast the discovery process as an optimization problem. Intuitively, we aim to find the parameter-sharing scheme that results in the best generalization capability. %
To estimate this generalization capability we use empirical data in a validation set. This results in a bi-level optimization:  optimize for the best parameter-sharing scheme  on a validation set, given that the model parameters which use this sharing are `optimal' on the training set.

\textbf{Contributions:}
\vspace{-0.15cm}
\begin{enumerate}
\item We theoretically analyze the benefits of learning a parameter-sharing scheme and show how to choose the validation and training set that are required in the proposed algorithm. For a family of multivariate Gaussian distributions with a shared mean, we show that the proposed method provably yields better generalization in terms of a mean squared error  than standard maximum likelihood training.

\item We also study how to evaluate the learned sharing scheme. Specifically, we advocate for the use of partition distance as a quantitative metric. This differs from  prior practice which relies on visual inspection.
Finally, we discuss practical considerations for using the proposed approach and validate its effectiveness through a range of experiments.  Empirically, we demonstrate that the approach can recover known permutation invariance and spatial equivariance from data.
\end{enumerate}

\section{RELATED WORK}\label{sec:rel}
We briefly highlight works on designing equivariance and invariance in machine learning. Then we review recent advances towards discovering equivariance from data. Lastly, we discuss how hyperparameter optimization is related to our work. 

{\noindent \bf Invariance and Equivariance.}
Designing invariance and equivariance representations has been widely utilized in building effective models. Well-known examples are hand-crafted features in computer vision, such as  SIFT~\cite{lowe1999object} which is scale invariant, or shift-invariant systems~\cite{vetterli2014foundations} in signal processing. Naturally, learning-based representations have also adopted these properties. For example, the widely used CNN~\cite{lecun1999object} or RNN~\cite{hochreiter1997long} are shift-invariant in space or time. The success of CNNs has also been generalized to other sets of equivariances. For example,~\citet{cohen2016group} propose a group-equivariant CNN, 
which is equivariant to rotations, reflections and translations, or TI-pooling~\cite{laptev2016ti}, which pools over the desirable transformations to achieve invariance. Other architectures, \eg, equivariant transformers~\cite{tai2019equivariant, fuchs2020se, romero2021group} have also been studied. 

Equivariance has also been extended to other domains, such as sets, which are permutation invariant~\cite{ravanbakhsh_sets, zaheer2017deep, qi2017pointnet, maron2020learning}, graphs~\cite{shuman2013emerging, defferrard2016convolutional, kipf2017semi, maron2018invariant}, meshes~\cite{dehaan2020gauge}, spherical images~\cite{cohen2018spherical, kondor2018clebsch}, key-points~\cite{YehNeurIPS2019}, trajectories~\cite{YehCVPR2019, LiuCORL2019, semtrack-2021} and tabular data~\cite{hartford2018deep}. These works demonstrate that designing equivariance into models/representations is beneficial. However, these approaches require a practitioner to select the suitable equivariance properties. Instead, in this work, we are interested in discovering this equivariance property explicitly from data rather than manually imposing it.

{\noindent \bf Learning Equivariance.} Recently,~\citet{benton2020learning} proposed to learn invariance for deep-nets from data. At a high-level, their approach achieves invariance by applying augmentations at the input and averaging the output. To learn the invariance, they parameterize the augmentation distribution and jointly learn these parameters with the deep-net's model parameters. Note that the model is only invariant to the sampled augmentations which may require many samples. \Eg, for the permutation group, their approach requires to sample all permutations to achieve invariance. In contrast, we achieve equivariance through parameter-sharing, and cast equivariance discovery as learning the parameter-sharing scheme.

In recent work,~\citet{zhou2020meta} consider learning equivariance in a meta-learning framework. Our work differs in the following ways: {(a)} We use an assignment matrix consisting of elements between zero and one to parameterize the sharing of parameters while~\citet{zhou2020meta} do not enforce any constraints. 
{(b)} This constraint permits to
quantitatively evaluated the discovered schemes. We propose to assess the discovered sharing scheme via the partition distance (PD)~\cite{gusfield2002partition}, while prior works rely on visual inspection.
{(c)} We demonstrate that the proposed method has advantages over standard maximum likelihood training without parameter-sharing on multivariate Gaussian distributions with a shared mean. We also show a trade-off between the size of training and validation sets.
Next, we  review hyperparameter optimization, as the sharing scheme can be viewed as a hyperparameter.

{\noindent \bf Hyperparameter Optimization.}
Typically formulated as a bi-level optimization problem, hyperparameter optimization consists of an upper/lower-level optimization task which, respectively, minimizes the loss on a validation/training set. Numerous hypergradient based methods have been proposed to solve this problem~\cite{Larsen, bengio2000gradient, maclaurin2015gradient, luketina2016scalable, Shaban, lorraine2019opt, RenYehNEURIPS2020}. %
\citet{lorraine2019opt} provide a comprehensive review. 

As bi-level optimization requires a validation set, here we also discuss how to select this set. Prior works have studied how to split a validation set~\cite{kearns1996bound, guyon1997scaling, amari1997asymptotic} and what test set size is necessary to yield a good generalization error estimate~\cite{guyon1998size}. More recently~\citet{afendras2019optimality}  study the optimal size of the validation set in the context of cross-validation. In this work, we optimize and study a specific hyperparameter,~\ie, the parameter-sharing scheme, to achieve equivariance discovery.
\section{PRELIMINARIES}\label{sec:prelim}
Abstractly,  equivariance and invariance capture properties of a function's input-output relationship. 
Consider the task of image segmentation. If an object is shifted within the image, one would expect the predicted segmentation to shift accordingly,~\ie, the model is \textit{shift-equivariant}. Similarly, for the task of image classification, the class prediction should remain the same for a shifted object. In this case, the classifier is \textit{shift-invariant}. Incorporating these properties into a multilayer perceptron (MLP) via parameter-sharing results in a convolutional neural net (CNN).
To generalize this success,~\citet{ravanbakhsh2017equivariance} theoretically study  types of equivariances that can be encoded via parameter-sharing, and how to construct such layers. We will briefly review their results in the remainder of this section. 

We start with the definitions of equivariance and invariance over the family of discrete group actions. 
A function $f: \mathbb{R}^{N} \mapsto \mathbb{R}^{M} $ is {\bf $\gG_{N, M}$-equivariant} if and only if~\textit{(iff)}
\bea
f(P_{\pi_N} \rvx) = P_{\pi_M}f(\rvx) \;\;\forall (\pi_N, \pi_M) \in \gG_{N, M},\; \rvx \in \mathbb{R}^{N},
\eea
where $P_\pi$ denotes a permutation matrix and $\gG_{N, M}$ denotes the set of group actions each characterized by two permutations $\pi_N$ and $\pi_M$ on input and output dimensions. Informally, equivariance describes how the output  changes when the input is transformed in a ``pre-defined way.''
Similarly, a function is {\bf $\gG_{N}$-invariant} \textit{iff}
\vspace{-0.4cm}
\bea
f(P_{\pi_N} \rvx) = f(\rvx) \;\;\forall \pi_N \in \gG_{N},\; \rvx \in \mathbb{R}^{N}.
\eea
Invariance is a special case of equivariance where $P_{\pi_M}=I_M$ is the identity matrix,~\ie, the output remains the same. Next, we discuss how to construct deep-nets that satisfy equivariance.

{\noindent\bf Equivariance Through Parameter-Sharing.} 
\citet{ravanbakhsh2017equivariance} theoretically study how to design deep-nets that are equivariant to any discrete group action, which are characterized above via permutation matrices. 
They prove that symmetries in model parameters, \ie, the sharing of the parameters, leads to equivariance. Importantly, a fully connected layer $f_{\mW}$ can be designed to be {$\gG_{N, M}$-equivariant} for any set of group actions. 
For example, a fully connected layer, $f_\mW(\rvx) \triangleq \mW\rvx$, where $\mW \in \mathbb{R}^{M \times N}$, is { $\gG_{N, M}$-equivariant} if the weight $\mW$ satisfies the following sharing scheme:
\bea
\mW_{m,n} = \mW_{\pi_M(m), \pi_N(n)}\;\; \forall(\pi_N, \pi_M) \in \gG_{N,M}.
\eea
They also propose a method to achieve such a sharing. 
In summary, by tying parameters, one can design fully connected layers that are equivariant to various discrete group actions. Note that~\citet{ravanbakhsh2017equivariance} \textit{design a model given an equivariance property},~\ie, a practitioner needs to decide which equivariance to build into a model. In contrast, we study \textit{how to discover equivariance} from data.

\vspace{-0.15cm}
\section{EQUIVARIANCE DISCOVERY BY LEARNED PARAMETER-SHARING}\label{sec:app}
\vspace{-0.15cm}
Here, we are interested in discovering equivariance from data, \ie, learning explicit parameter-sharing schemes rather than manually imposing them. 

Consider a supervised learning setup, given a dataset $\cD=\{(\rvx, \rvy)\}$, the goal is to learn the parameters $\bm\theta$ of a model $f_{\bm\theta}(\rvx)$, %
by minimizing a desired loss function $\cL$, \ie,  %
\begin{equation}\label{eq:sl}
\min_{\bm\theta} \cL(\bm\theta, \cD) = \min_{\bm\theta} \sum_{(\rvx,\rvy) \in \cD} \ell(f_{\bm\theta}(\rvx), \rvy).
\end{equation}
To discover equivariance, we introduce a parametric representation of the sharing scheme and develop an algorithm to optimize over it. 

{\noindent\bf Parameterizing Parameter-Sharing.} We  use an assignment matrix $\mA$ to select which of the parameters are shared. Formally, let %
\begin{equation}
\bm\theta = \mA\bm\psi,
\end{equation}
where $\bm\theta, \bm\psi \in \mathbb{R}^{K}$, $\mA \in \{0,1\}^{K \times K}$, and $\mA$ is %
row stochastic, \ie, $\forall i$ $\sum_j \mA_{ij} = 1$. Hence, entries in $\bm\theta$ may originate from the same entries in $\bm\psi$. Using $\mA$, we can represent all possible sharing configurations, which in turn permits to incorporate different equivariances. For example, $\mA$ encoding a Toeplitz matrix results in the convolution operation, \ie, shift equivariance. With this parametrization at hand, we now describe how to learn both $\mA$ and $\bm\psi$ from data.

{\noindent \bf Learning Parameter-Sharing.} 
Note, jointly optimizing $\bm\psi$ and $\mA$ on $\cD$ doesn't yield the desired result: 
the trivial solution  $\mA=\mI$ 
selects all the parameters, leading to the lowest  loss on the training set. This is not necessarily desirable. 

Recall, the motivation for an equivariant model is to improve {\it generalization}. Therefore, we directly estimate  generalization  on data. For this we split the dataset $\cD$ into two sets, the training set $\gT$ and the validation set $\gV$. We then aim to solve the following bi-level program:
\bea
\overbracket{\min_\mA \cL(\underbrace{\mA\bm\psi^*(\mA)}_{\bm\theta}, \cV)}^{\text{upper-level task}}
\nonumber
\text{~s.t.~} \bm\psi^*(\mA)=\overbracket{\argmin_{\bm\psi}
\cL(\mA\bm\psi, \gT)}^{\text{lower-level task}},
\eea
\vspace{-0.7cm}
\bea\label{eq:struct_main}
\hspace{2cm}~\mA \in \{0,1\}^{K \times K}, \sum_j \mA_{ij} = 1~~\forall i.
\eea
Intuitively, we aim to find the ``best sharing scheme'' on validation set $\gV$ (upper-level task), given model parameters $\bm\psi$  trained on $\gT$ (lower-level task). Once having discovered the optimal parameter-sharing scheme $\mA_{\tt val}$, we fix it and train $\bm\psi$ on the entire dataset $\cD$ to obtain the best model.

This naturally leads to the following questions: 
(a) Is learning $\mA$ beneficial?; and 
(b) Can the optimization in~\equref{eq:struct_main} be solved efficiently?
We analyze (a)  in~\secref{sec:analyze}, followed by discussing considerations for (b) in~\secref{sec:compute}.

\subsection{Analysis on Gaussian Data}\label{sec:analyze}
We analyze the approach given in \equref{eq:struct_main} on $K$-dimensional Gaussian vectors that  are independent and identically distributed (\textit{i.i.d.}), \ie,
\be\label{eq:data_gauss}
\rvy \sim \cN(\mA_{\tt gt}{\bm{\psi}_{\tt gt}}, \sigma^2\mI).
\ee
Here, $\mA_{\tt gt}$ denotes the unknown ground-truth sharing scheme. The task is to estimate the ground-truth mean, 
${\bm \theta_{\tt gt}} \triangleq \mA_{\tt gt}\bm\psi_{\tt gt}$ and the sharing scheme $\mA_{\tt gt}$. With our supervised learning setup,  $\cD = \{\rvy\}$, $f_{\bm\theta} = \bm\theta$ and $\ell = \ell_2$.

{\noindent\bf Benefits of Learning $\mA$.}
Let $\hat{\bm\theta}(\gD)$  be the maximum likelihood estimate on dataset $\gD$. To analyze, we consider the mean squared error (MSE)  of an estimator, 
\bea\label{eq:mse}
\text{MSE}(\hat{\bm\theta}(\gD)) \triangleq \mathbb{E}\norm{\hat{\bm\theta}(\gD)-\bm\theta_{\tt gt}}^2\\\nonumber
 = \norm{\text{Bias}(\hat{\bm\theta}(\gD))}^2 + \text{Trace}(\mathbb{V}(\hat{\bm\theta}(\gD))),
\eea
where $\text{Bias}(\hat{\bm\theta}) = \mathbb{E}(\hat{\bm\theta})-\bm\theta_{\tt gt}$, 
and $\mathbb{V}(\hat{\bm\theta})$ denotes the covariance matrix of $\hat{\bm\theta}$. {\bf Note:} the expectation is with respect to the distribution that generated the finite dataset with $|\cD|$ number of samples, \ie, $\hat{\bm\theta}(\gD)$ is a random variable and $\bm\theta_{\tt gt}$ is fixed~\cite{wasserman2013all}. 

We further let  $\mA_{\tt val}$ denote the sharing scheme discovered using~\equref{eq:struct_main} and $\hat{\bm \theta}_{\tt val}(\gD)$ is the maximum likelihood estimate on $\gD$ following the sharing scheme $\mA_{\tt val}$, \ie, $\hat{\bm\theta}_{\tt val} =\mA_{\tt val}\bm\psi_{\tt val}$. %
With the notation defined, we study the form of the mean squared error given an estimator using the parameter-sharing scheme $\mA_{\tt val}$. Specifically, we identify the role of the rank  $\Rank{\mA_{\tt val}}$ in the MSE.
\vspace{0.07cm}

\begin{mdframed}[style=MyFrame,align=center]
\vspace{0.1cm}
\begin{restatable}{claim}{vb}{
For Gaussian data (\equref{eq:data_gauss}) 
and a given sharing scheme $\mA_{\tt val}$ we have\vspace{-0.2cm}$$
\text{MSE}(\hat{\bm\theta}_{\tt val}(\gD))=\!\norm{\mA_{\tt val}\bar{\mA}^{\intercal}_{\tt val}\bm\theta_{\tt gt} - \bm\theta_{\tt gt}}^2 + \frac{\Rank{\mA_{\tt val}}\sigma^2}{|\gD|},
$$
where $\bar{\mA}_{\tt val}$ refers to the column normalized $\mA_{\tt val}$ and $\Rank{\cdot}$ denotes the rank of a matrix.
}
\label{clm:vb}
\end{restatable}
\vspace{-0.1cm}
\end{mdframed}
\begin{proof}
\vspace{-0.2cm}
Let $S_i$ denote the set of indices that share parameters for the $i^{\text{th}}$ dimension of $\hat{\bm\theta}_{\tt val}$ as characterized %
in $\mA_{\tt val}$. 
Formally, $S_i \triangleq \{k \in \{1,\hdots, K\}|\; \forall j\; \mA_{\tt val}[i,j]=1 \land \mA_{\tt val}[k,j]=1\}$. 
As the dimensions and samples are independent, the maximum likelihood estimator is the average over the shared dimensions and samples, \ie,
\be \label{eq:bias_sub}
\mathbb{E}(\hat{\bm\theta}_{\tt val}[i]) = \frac{1}{|S_i|}\sum_{k \in S_i} {\bm \theta_{\tt gt}[k]},
\ee
Similarly, the variance is
\be \label{eq:var_sub}
\mathbb{V}(\hat{\bm \theta}_{\tt val}[i]) = \frac{\sigma^2}{|S_i||\gD|}.
\ee
Substituting~\equref{eq:bias_sub} and~\equref{eq:var_sub} into the MSE definition in~\equref{eq:mse} concludes the proof. Additional details are deferred to Appendix~\secref{sec:supp_vb_proof}.
\end{proof}

Consider the case without any parameter-sharing, \ie, the parameters are \underline{ind}ependent ($\mA_{\tt ind}$ = $\mI$). As the estimator is unbiased, we obtain 
\be
\text{MSE}(\hat{\bm\theta}_{\tt ind}(\gD)) = \frac{K\sigma^2}{|\gD|}.
\ee
Observe that when $K \geq \Rank{\mA_{\tt val}} \geq \Rank{\mA_{\tt gt}}$ %
then there exists an \emph{unbiased} $\mA_{\tt val}$ (as rank is larger) such that the MSE (in \clmref{clm:vb}) is lower than the MSE without parameter sharing (\ie, when using $\mA_{\tt ind}$). 
Specifically, there exists an unbiased estimator, $\text{Bias}(\hat{\bm\theta})=0$, with a lower variance term compare to the estimator $\hat{\bm\theta}_{\tt ind}$ without any parameter sharing.
This means that it is possible to find an estimator that generalizes better in terms of MSE.

Next, we show that the procedure in~\equref{eq:struct_main} will select such an $\mA_{\tt val}$. 
Recall, the algorithm selects $\mA_{\tt val}$ based on the loss over the empirically sampled validation set $\cV$, \ie, 
\be\label{eq:l2_loss}
\cL(\hat{\bm\theta}, \gV) = \sum_{\rvy \in \gV} \norm{\hat{\bm\theta}-\rvy}^2 \geq \norm{\hat{\bm\theta} - \frac{1}{|\gV|}\sum_{\rvy \in \gV} \rvy}^2.
\ee
\equref{eq:l2_loss} (left-hand side) is an upper bound on  the squared error between the parameter estimates.  Left- and right-hand side have the same global minimum. Therefore, we directly analyze $\widehat{\text{MSE}}$, an MSE estimate based on the validation set $\gV$, defined as follows:
\be
\widehat{\text{MSE}}(\hat{\bm\theta}) = \mathbb{E}\norm{\hat{\bm \theta}- \hat{\bm \theta}_{\tt ind}(\cV)}^2.
\ee
Note that the expectation is \wrt  the estimator $\hat{\bm \theta}$.

Recall, the procedure in~\equref{eq:struct_main} selects the sharing scheme $\mA_{\tt val}$ which minimizes the $\widehat{\text{MSE}}$. By law of large numbers, when $|\cV| \rightarrow \infty$ then $\widehat{\text{MSE}}(\hat{\bm\theta}) \rightarrow \text{MSE}(\hat{\bm\theta})$. This means that the program in~\equref{eq:struct_main} characterizes the $\mA_{\tt val}$ which minimizes the MSE if $\gV$ is sufficiently large. 

In practice, we have a limited amount of data. Therefore we further study the finite sample behavior and how to decide the sizes of training and validation sets.

{\noindent\bf Finite Sample Analysis.} We study the MSE gap 
\bea\label{eq:mse_gap}
\MSE{\hat{\bm\theta}_{\tt val} (\gD)} - \MSE{\hat{\bm\theta}_{\tt gt} (\gD)}
\eea
between learned and ground-truth \textit{sharing scheme}. This gap is useful as it  measures the quality of the estimator of the proposed procedure. Specifically, we construct an upper bound to identify the role of  dataset sizes $|\gT|$ and $|\gV|$. Recall,  that we split a given dataset $\gD$ into a training set $\gT$ and a validation set $\gV$, which affects the characterized $\mA_{\tt val}$.

\vspace{0.15cm}
\begin{mdframed}[style=MyFrame]%
\vspace{0.1cm}
\begin{restatable}{claim}{bound}{
Given data drawn \textit{i.i.d.} following~\equref{eq:data_gauss}, with probability $1-\alpha$ and $\alpha < \exp{\frac{-K}{10}}$,  the MSE gap in~\equref{eq:mse_gap} is upper bounded by
\vspace{-0.2cm}
\bea\label{eq:bound}
\sigma^2 \Big( \underbrace{\frac{1-r}{r|\gD|} \left(\Rank{\mA_{\tt gt}}-1\right)}_{\text{sharing rel.\ MSE gap}} -  \underbrace{\frac{40\ln(\alpha)}{(1-r)|\gD|}}_{\text{con.\ rel.\ MSE gap}} \Big),
\eea
where $r = \frac{|\gT|}{|\gD|}$ denotes the ratio between the size of training and overall dataset.
}
\label{clm:bound}
\end{restatable}
\vspace{-0.1cm}
\end{mdframed}
\vspace{-0.3cm}
\begin{proof}[Proof sketch]
The high-level idea is to decompose~\equref{eq:mse_gap} into three parts:
\bea
\label{eq:mse_gap1}
 & &\MSE{\hat{\bm \theta}_{\tt val}(\gD)} -  \MSE{\hat{\bm \theta}_{\tt val}(\gT)},\\
\label{eq:mse_gap2}
+& &\MSE{\hat{\bm \theta}_{\tt val}(\gT)} -  \MSE{\hat{\bm \theta}_{\tt gt}(\gT)},\\
\label{eq:mse_gap3}
+&  &\MSE{\hat{\bm \theta}_{\tt gt}(\gT)} -  \MSE{\hat{\bm\theta}_{\tt gt} (\gD)}.
\eea
We prove the claim by upper bounding each of the terms. 
The complete proof is deferred to Appendix~\secref{sec:supp_gauss_proof}. 
\end{proof}

\begin{figure}[t]
\centering
\includegraphics[width=.85\linewidth]{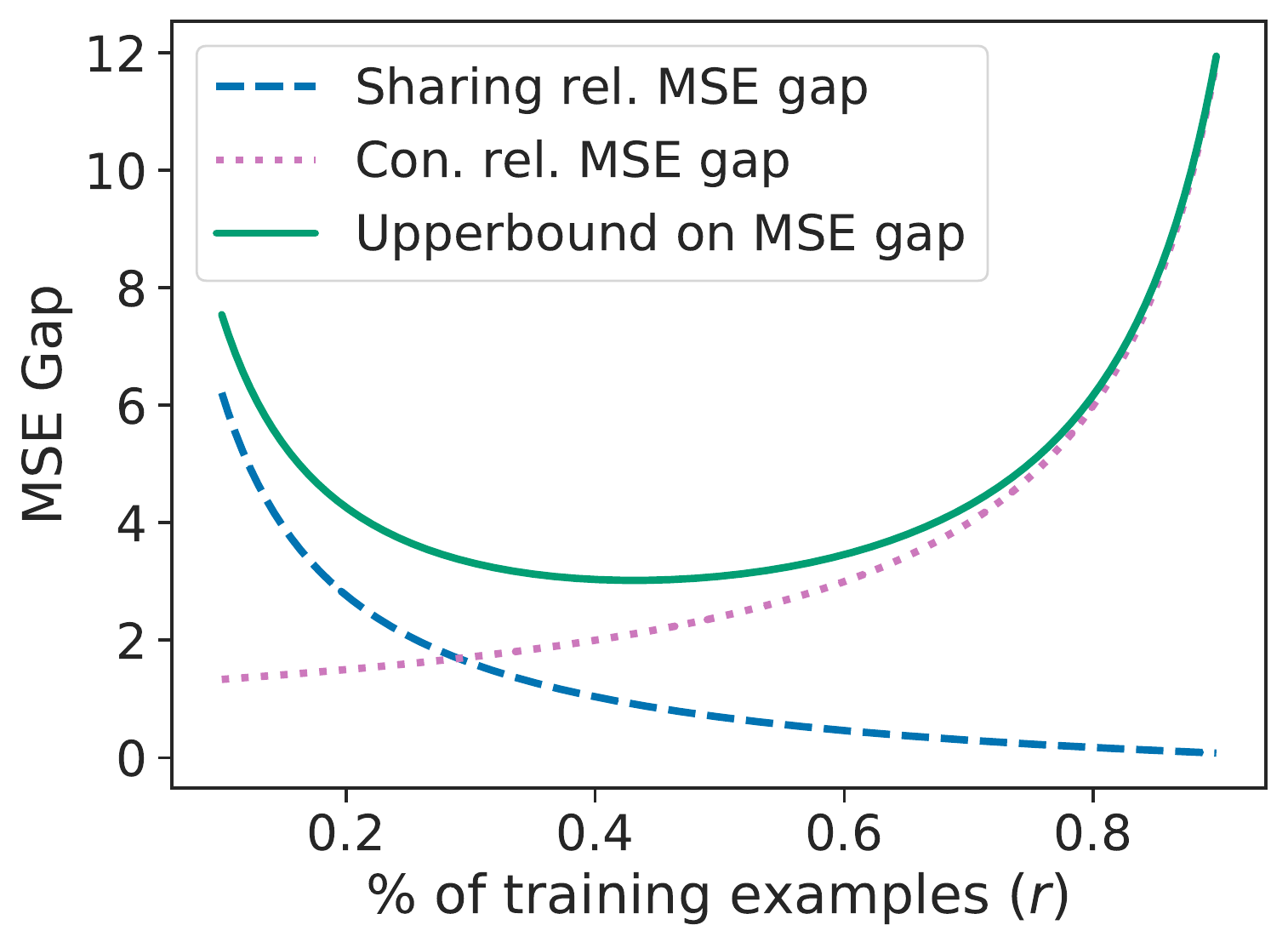}
\vspace{-0.3cm}
\caption{Illustration of the upper bound on the MSE gap in~\equref{eq:bound}.
\vspace{-0.1cm}
}
\label{fig:bound_viz}
\end{figure}

We now highlight this results' significance.
Observe that the upper bound in~\equref{eq:bound} consists of two terms, a \textit{sharing} related MSE gap and a \textit{confidence} related MSE gap. The two terms form a trade-off \wrt the optimal percentage of the training examples. We provide an illustration in~\figref{fig:bound_viz} demonstrating the trade-off between the sharing term and the confidence term. Note, the optimal validation set size may be much larger than the commonly used 80-20 train/val split.

Intuitively, when there is less parameter sharing, \ie, $\Rank{\mA_{\tt gt}}$ is large, then more training data should be used to get a good estimate of the model parameters. Similarly, if one aims to have more confidence in $\mA_{\tt val}$, \ie, $-\ln(\alpha)$ is large, then more validation data should be used. 
Also, this bound suggests that it can be desirable to use a validation set that is larger than the training set, 
which is not a common practice to date. 
 
Further, we can use the upper bound to  identify data distributions where the proposed algorithm is provably better in generalization than standard maximum likelihood training without parameter sharing. 

For example, when $\Rank{\mA_{\tt gt}}=1$, 
\bea\nonumber
\MSE{\hat{\bm\theta}_{\tt val}(\cD)} &\leq& \frac{-40\ln\alpha}{(1-r)|\cD|}\sigma^2 + \MSE{\hat{\bm\theta}_{\tt gt}(\cD)}\\\nonumber
&=& \frac{-40\ln\alpha}{(1-r)|\cD|}\sigma^2 + \frac{\sigma^2}{|\cD|}\\
&\leq& \frac{K\sigma^2}{|\cD|} = \MSE{\hat{\bm\theta}_{\tt ind}}
\eea
given that $K$ is sufficiently large. 
This shows that the sharing approach achieves a lower MSE than standard maximum likelihood training where all the parameters are independent, \ie, $\hat{\bm \theta}_{\tt ind}$. 

We will next discuss how to address the bi-level optimization in~\equref{eq:struct_main}.

\subsection{Practical Considerations}\label{sec:compute}
\begin{figure}[t]
\vspace{-.3cm}
\begin{minipage}{0.49\textwidth}
\begin{algorithm}[H]
\begin{algorithmic}[1]
\STATE Initialize model parameters $\bm\psi$, $\mA$ %
\WHILE {not converged} 
\STATE Sample batch $\gV' \subseteq \gV$.
\STATE {\color{CadetBlue} \# Solve lower-level task.}
\STATE $\bm\psi^*(\mA) \leftarrow \arg\min_{\bm \psi} \cL(\mA\bm \psi, \gT)$
\STATE $\mA \leftarrow \mA - \eta \cdot \nabla_\mA \big(\cL(\mA\bm\psi^*,\gV')+H(\mA) + \norm{\mA}_* \big)$
\ENDWHILE 
\STATE $\mA_{\tt val} \leftarrow \mA$
\STATE {\color{CadetBlue} \# Train with all the data with fixed $\mA_{\tt val}$.}
\STATE $\bm\psi^*(\mA_{\tt val}) \leftarrow \arg\min_{\bm \psi} \cL(\mA_{\tt val}\bm \psi, \gD)$
\STATE {\bf Return} $\mA_{\tt val}, \bm\psi^*(\mA_{\tt val})$
\end{algorithmic}
\caption{Equivariance discovery via learned parameter-sharing}
\label{alg:training}
\end{algorithm}
\end{minipage}
\end{figure}
For really small-scale problems a brute-force search over all $\mA$ solves the proposed program in \equref{eq:struct_main}. However, brute-force search quickly becomes infeasible  when the dimensions grow. 
Instead of brute-force search, we relax the integrality constraint to $\mA \in [0,1]^{K \times K}$. This permits the use of continuous optimization, \eg, projected gradient descent on $\mA$ or use of a softmax to avoid the constraints altogether. 

However, this relaxation may yield a result  that doesn't satisfy the original constraints. To alleviate this scenario, we found use of two penalty functions to help: 
\bea
H(\mA) = -\sum_{i,j} \log(\mA[i,j]) \cdot \mA[i,j] \\ \text{~and~}\norm{\mA}_* = \text{trace} \left(\sqrt{\mA^{\intercal}\mA} \right).
\eea
The first entropy term encourages elements of $\mA$ to be closer to 0 or 1. The second nuclear norm encourages $\mA$ to be low-rank.
Empirically we find both to improve robustness to  random initializations of the model parameters. We illustrate the overall algorithm in~\algref{alg:training}, where we iteratively solve the upper-level optimization via mini-batch gradient descent. In practice, we monitor the validation loss to determine convergence. Additionally, more advanced gradient based optimization algorithms, \eg, Adam~\cite{kingma2015adam}, can be utilized.
A natural question is how to quantitatively evaluate the recovered sharing scheme, which we discuss next.

\subsection{Quantitative Evaluation of Equivariance}\label{sec:quan_eval}
Prior works rely on visual inspection of the parameter sharing to assess the quality, or use the final task performance as a surrogate; both of which are not a direct comparison with the ground-truth sharing scheme. The main challenge is that an element-wise distance between $\mA_{\tt val}$ and $\mA_{\tt gt }$ is not meaningful, as $\mA$ is unique up-to permutations. 

Hence, we propose to use the {\it Partition Distance (PD)}~\citep{gusfield2002partition} as an evaluation metric. Specifically, PD between two sharing schemes measures the number of assignments that must be changed for one sharing scheme to be identical to the other. We will show that PD is related to the symmetric difference of the equivariance groups $\gG$ encoded by sharing scheme $\mA$. 
We first review the definitions of cluster, partition, and  partition distance \citep{gusfield2002partition}.

\vspace{0.1cm}
\begin{mdframed}[style=MyFrame]%
\vspace{0.1cm}
\begin{definition}[Cluster]
Given a set $\gS$, a cluster is a non-empty subset of $\gS$, \ie, $\gC \subseteq \gS$ where $\gC \neq \emptyset$.  
\end{definition}

\begin{definition}[Partition]
A partition of $\gS$ is a set of clusters, $\gP_\gS = \{\gC_i\}$, where $\gC_j \cap \gC_k = \emptyset \;\;\forall j \neq k$ and $\bigcup_{i=1}^{|\gP_\gS|} \gC_i = \gS$. In other words, the elements in $\gS$ are ``partitioned'' into mutually exclusive sets.
\end{definition}

\begin{definition}[Partition Distance]
Given two partitions of a set, the partition distance is \underline{the number of elements that must be moved} between  clusters such that the two partitions are identical.
\end{definition}
\vspace{-0.1cm}
\end{mdframed}

Next, consider the Gaussian with shared means problem in~\secref{sec:analyze} where we model the mean as
$\bm\theta = \mA\bm\psi.$
Here, $\bm\theta, \bm\psi \in \mathbb{R}^{K}$, $\mA \in \{0,1\}^{K \times K}$, and $\mA$ is row stochastic, \ie, $\forall i$ $\sum_j \mA_{ij} = 1$. 
A parameter-sharing scheme $\mA$ can be viewed as a partition over the set of model parameters. As $\mA$ is integral row stochastic, %
it forms a partition of mutually exclusive clusters. 

Given two parameter-sharing schemes $\mA^{(1)}$ and $\mA^{(2)}$ the parition distance $PD(\mA^{(1)}, \mA^{(2)})$ can be efficiently computed, in polynomial time, as proposed by~\citet{gusfield2002partition}. We will next explain how this distance relates to equivariance. 

For a sharing scheme $\mA^{(i)}$, we first construct sets to form a partition $\gP_K^{(i)}$ consisting of clusters $\gC^{(i)}_k$ indicating indices of shared parameters, \ie, 
\bea
\gP_K^{(i)} &=& \{ \gC^{(i)}_k | \;\forall k \in [1,\dots, K]\} \text{ and }\\ \gC^{(i)}_k &=& \{j  | \mA^{(i)}[j,k]=1\}.
\eea

Due to the shared parameters, the indices within a cluster %
can be permuted. \Ie, given $\mA^{(i)}$, the model is $\gG_{K,K}^{(i)}$-equivariant, where 
\be
\gG_{K,K}^{(i)} = \bigcup_{\gC^{(i)}_k \in \gP^{(i)}_k} \Pi_{\gC^{(i)}_k}.
\ee
We use $\Pi_{\gC^{(i)}_k}$ to denote the set of all possible permutation matrices over the indices in $\gC^{(i)}_k$, while holding the other indices fixed. %

For an effective evaluation metric, it should capture the similarity between two equivariances, $\gG^{(1)}$ and $\gG^{(2)}$, each  encoded by respective sharing schemes, $\mA^{(1)}$ and $\mA^{(2)}$.
We consider the symmetric difference of two sets to quantify the similarity between $\gG$-equivariances, \ie, 
\be
\gG^{(1)} \Delta \gG^{(2)} = (\gG^{(1)} - \gG^{(2)}) \cup  (\gG^{(2)} - \gG^{(1)}).
\ee
This captures the non-overlapping elements within groups. We now relate this quantity to the PD.

\vspace{0.2cm}
\begin{mdframed}[style=MyFrame]
\vspace{0.1cm}
\begin{restatable}{claim}{pd}{
Considering the Gaussian sharing setup,
\bea\nonumber
\kappa PD(\mA^{(1)}, \mA^{(2)}) \geq |\gG_{K,K}^{(1)} \Delta \gG_{K,K}^{(2)} | \geq PD(\mA^{(1)}, \mA^{(2)}),
\eea
where $\kappa = K!-1$ is a constant and $K$ is the number of dimensions.
}\label{clm:pd}
\end{restatable}
\end{mdframed}
\begin{proof} 
Deferred to~\secref{sec:supp_pd_proof} in the Appendix.
\end{proof}

From~\clmref{clm:pd}'s upperbound, 
\be
PD(\mA_1, \mA_2)=0 \;\;\longrightarrow\;\; |\gG_{N,N}^{(1)} \Delta \gG_{N,N}^{(2)} |=0.
\ee 
Hence, when a partition distance is zero, the two sharing schemes achieve the same equivariance. Next, as the symmetric difference is lower bounded by the partition distance, when the PD is non-zero, then the two equivariances are not identical. Base on these properties and its efficiency to compute, we find partition distance to be a suitable evaluation metric.
\vspace{-0.25cm}

\begin{figure*}[t]\centering
\begin{minipage}[b]{0.49\textwidth}
\includegraphics[height=0.36\textwidth, width=0.48\textwidth]{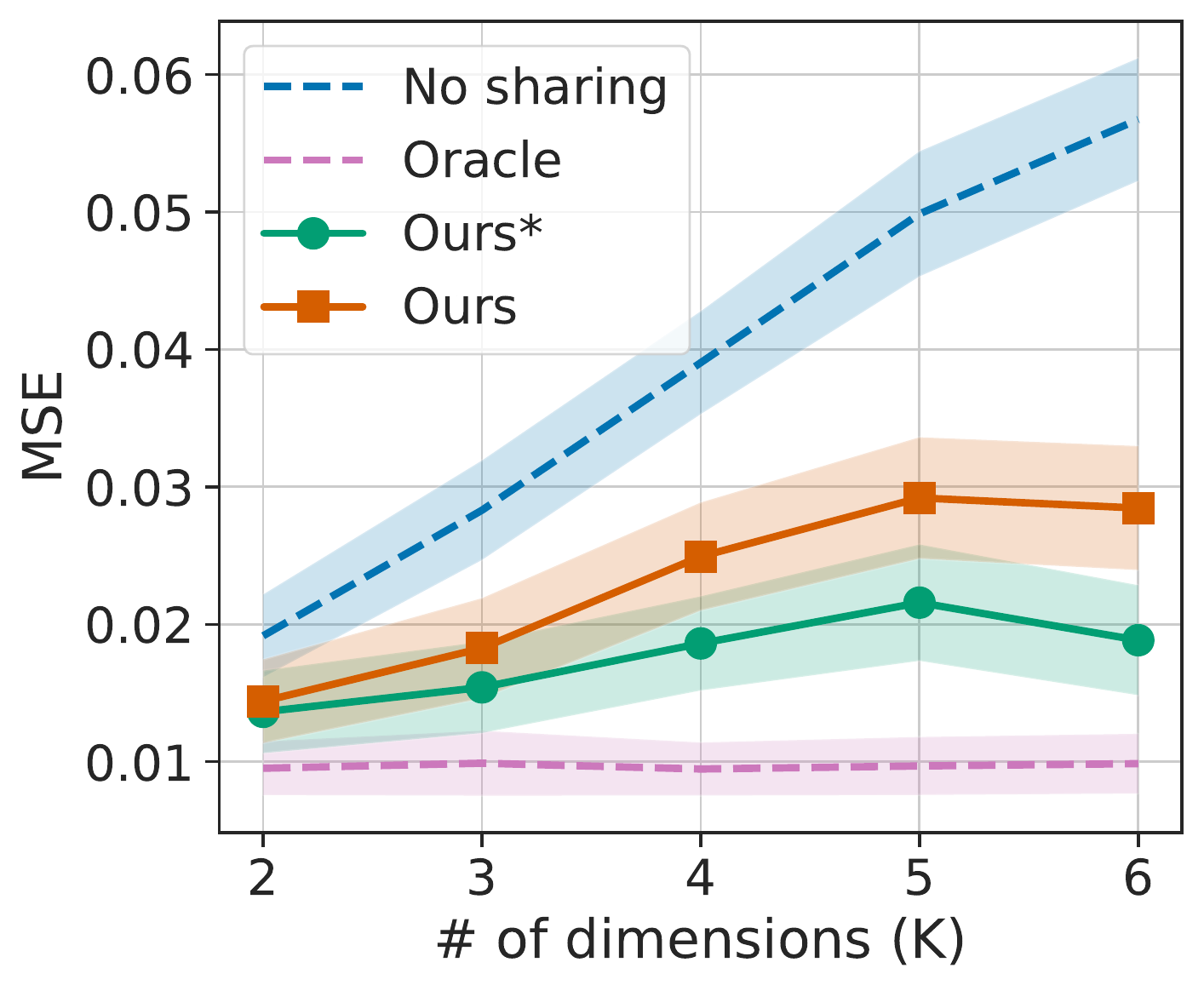}
\includegraphics[height=0.36\textwidth, width=0.48\textwidth]{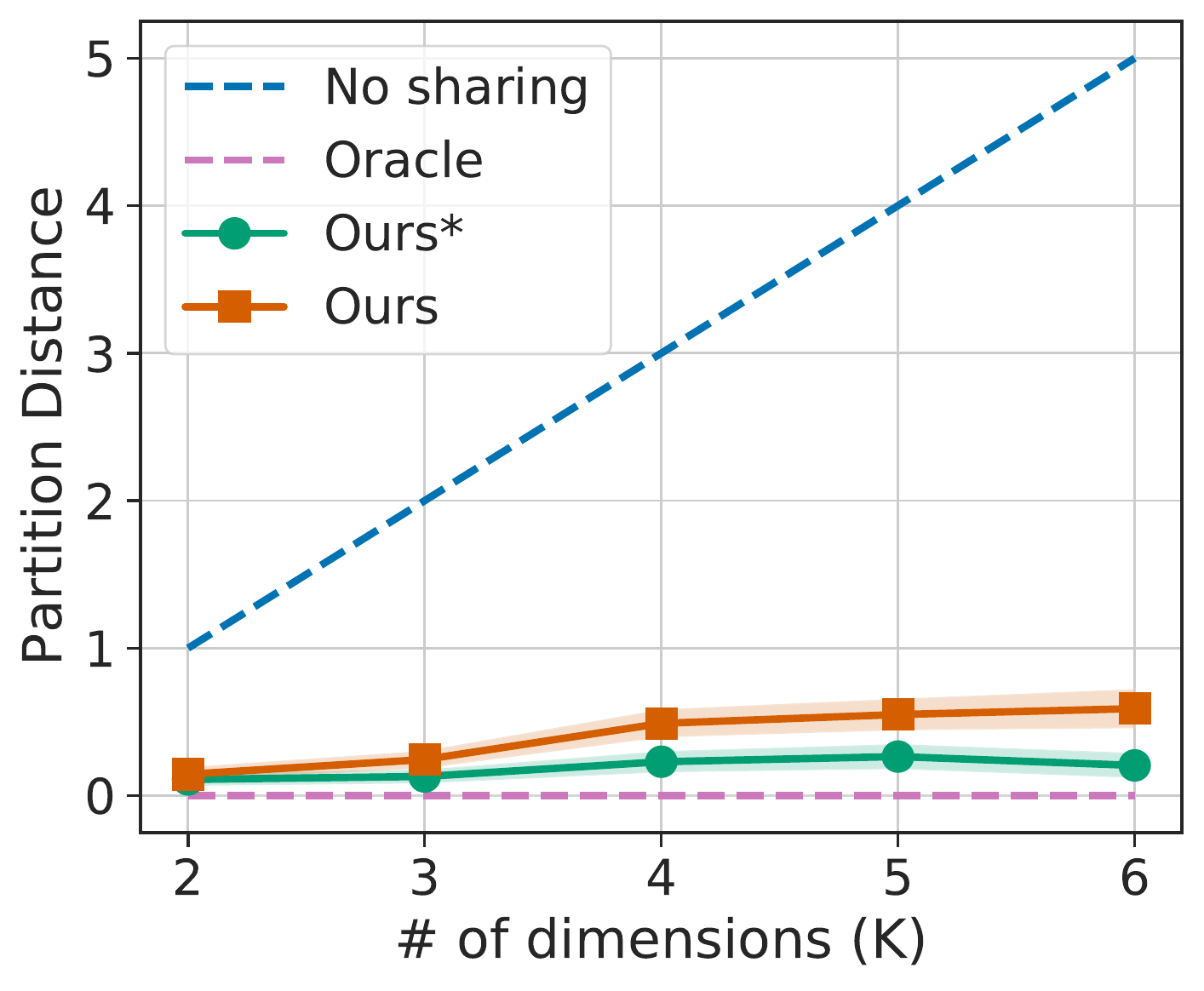}
\vspace{-0.3cm}
\caption{
MSE/PD \vs \# of dimensions.}
\label{fig:quan_ra1}
\end{minipage}
\begin{minipage}[b]{0.49\textwidth}
\includegraphics[height=0.36\textwidth, width=0.48\textwidth]{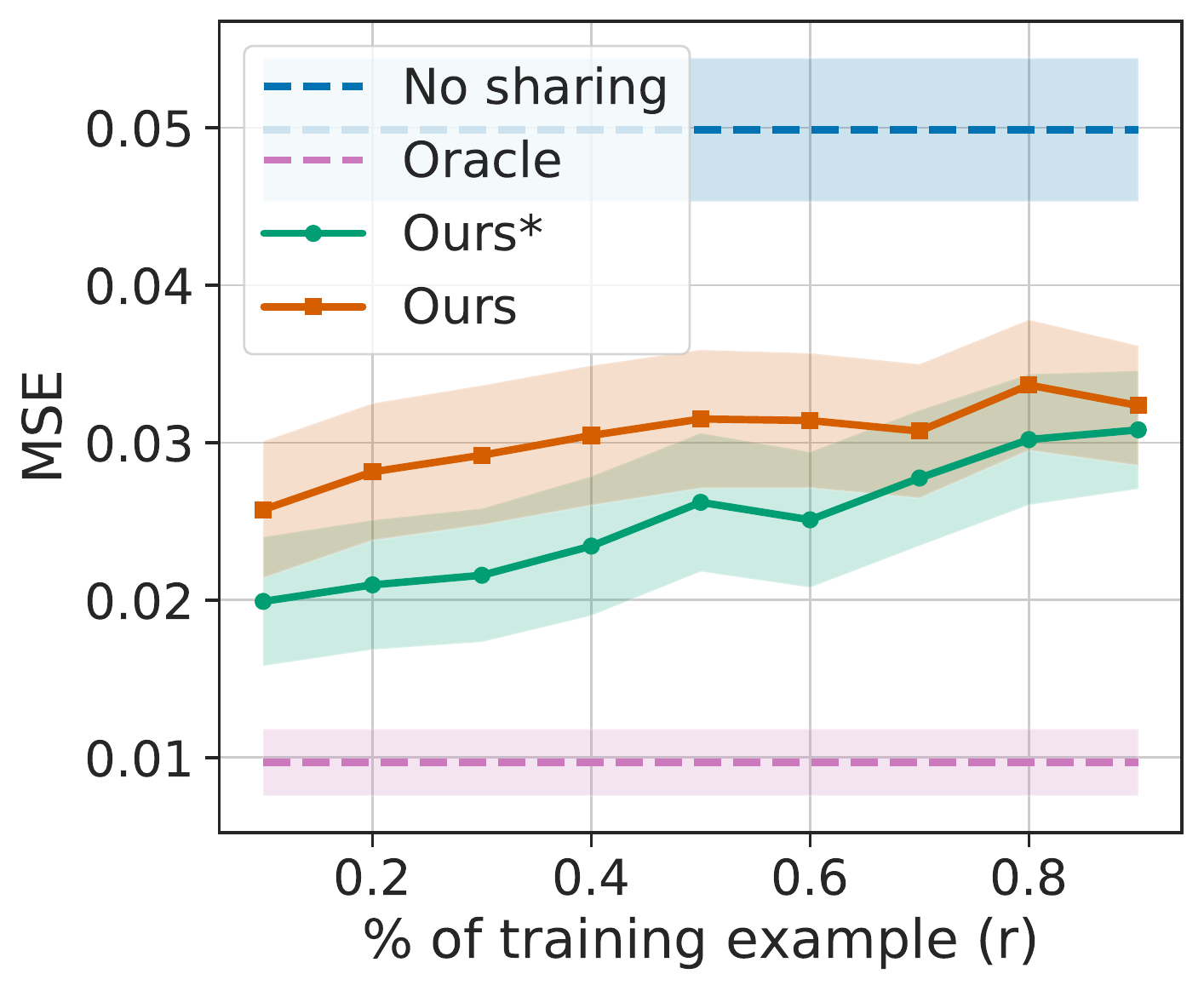}
\includegraphics[height=0.36\textwidth, width=0.48\textwidth]{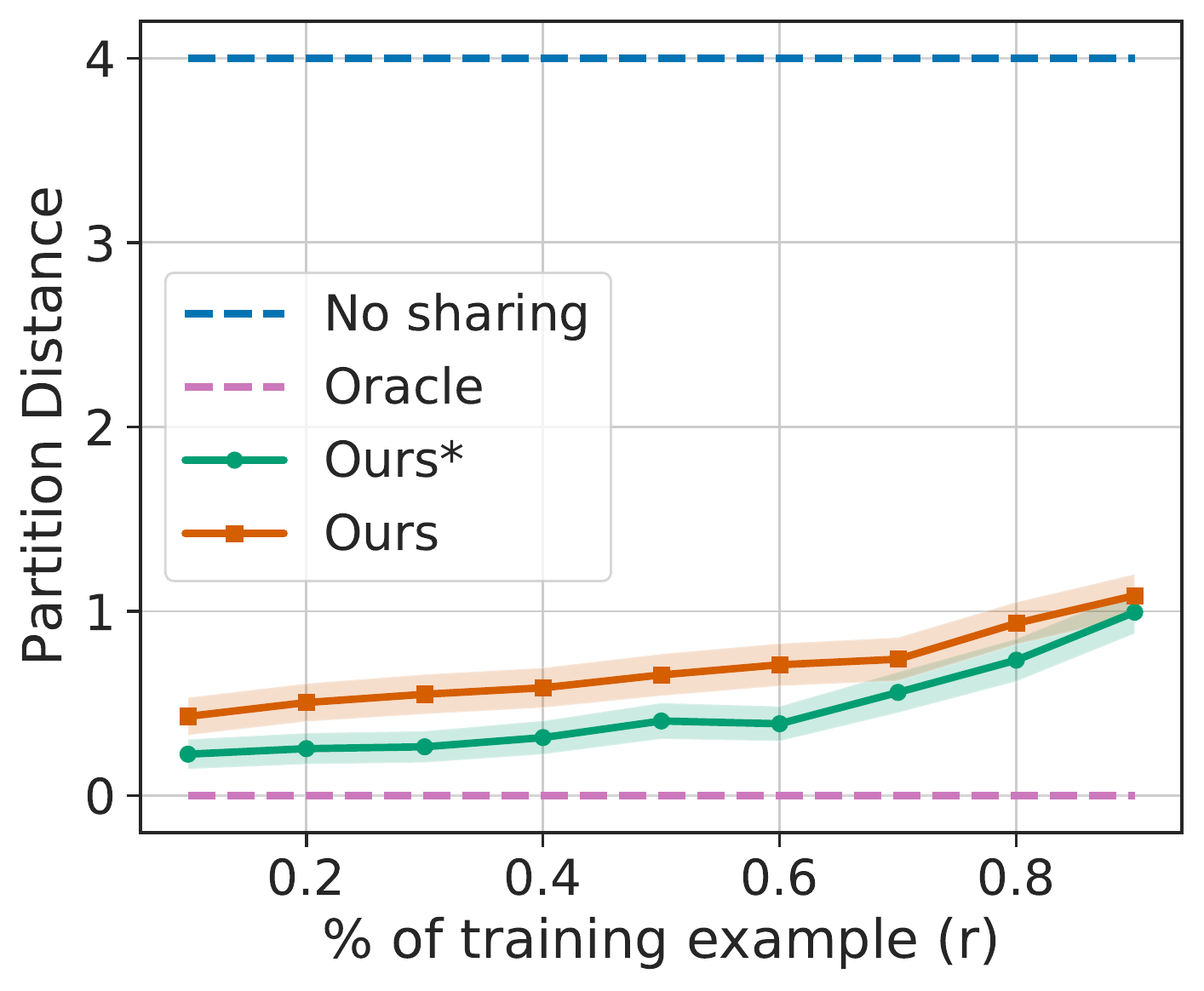}
\vspace{-0.3cm}
\caption{MSE/PD \vs \% of training data.}
\label{fig:quan_ra2}
\end{minipage}\vspace{0.2cm}
\begin{minipage}[b]{0.49\textwidth}
\includegraphics[height=0.36\textwidth, width=0.48\textwidth]{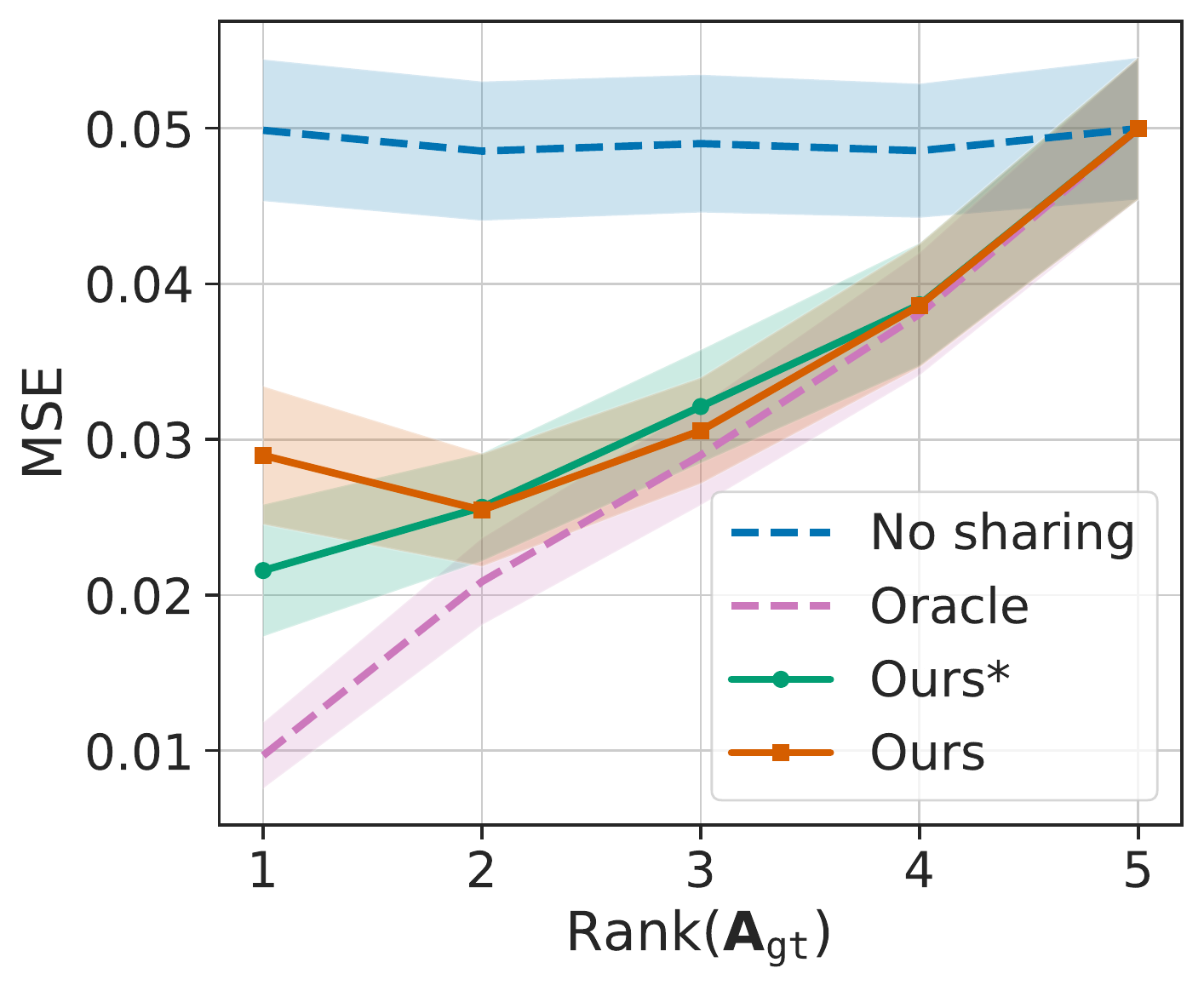}
\includegraphics[height=0.36\textwidth, width=0.48\textwidth]{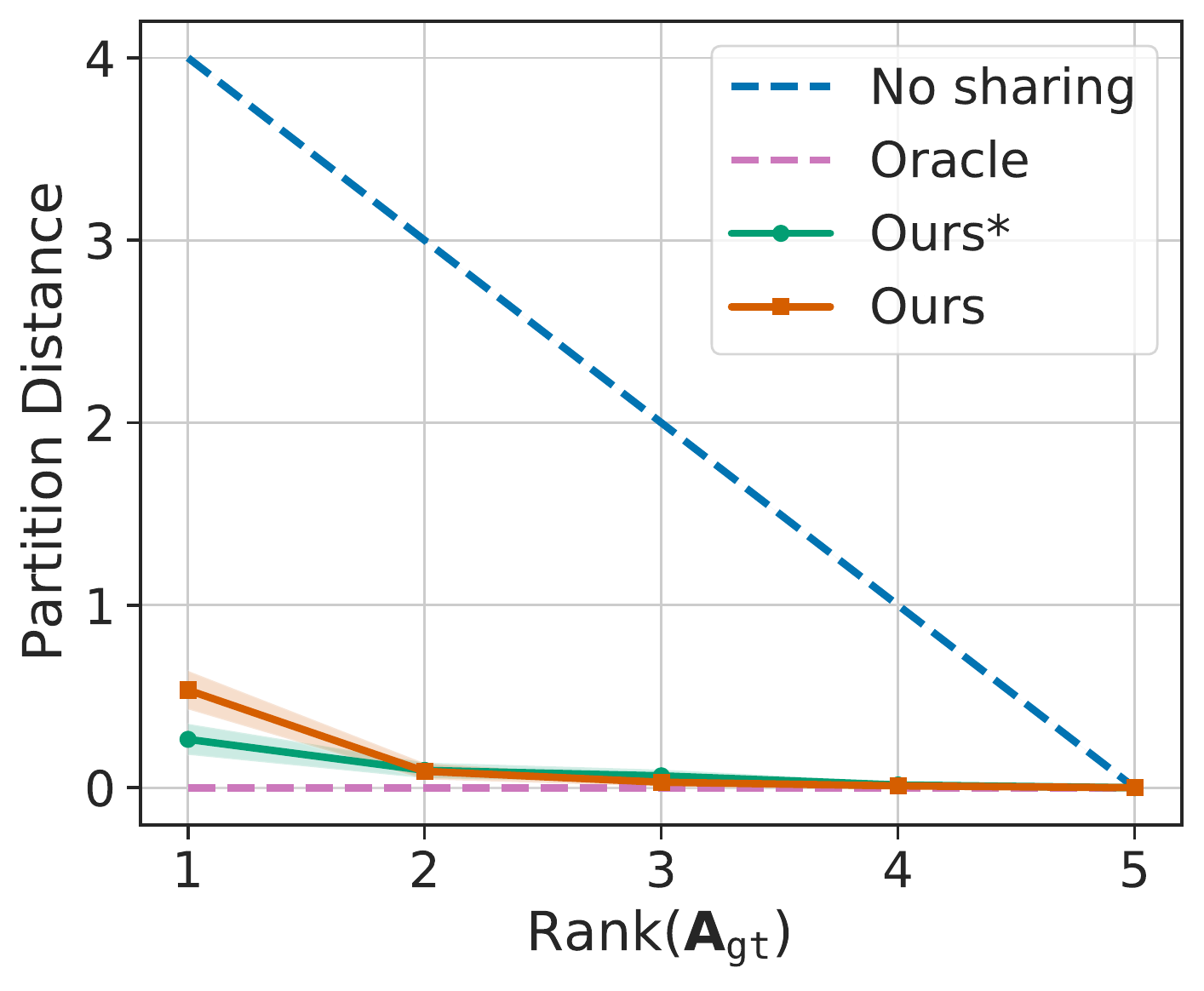}
\vspace{-0.3cm}
\caption{MSE/PD \vs \Rank{$\mA_{\tt gt}$}.} 
\label{fig:quan_ra3}
\end{minipage}
\begin{minipage}[b]{0.49\textwidth}
\includegraphics[height=0.36\textwidth, width=0.48\textwidth]{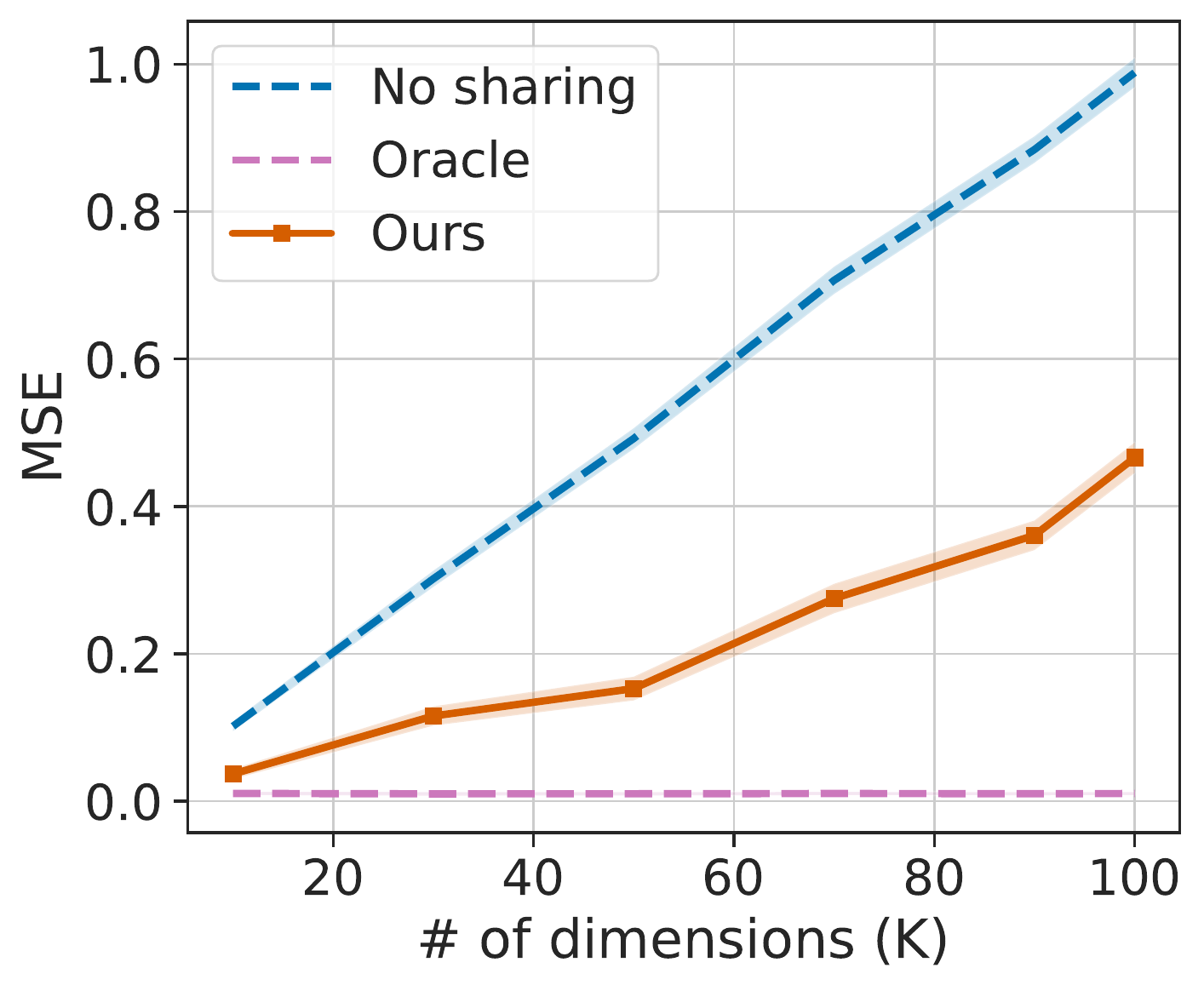}
\includegraphics[height=0.36\textwidth, width=0.48\textwidth]{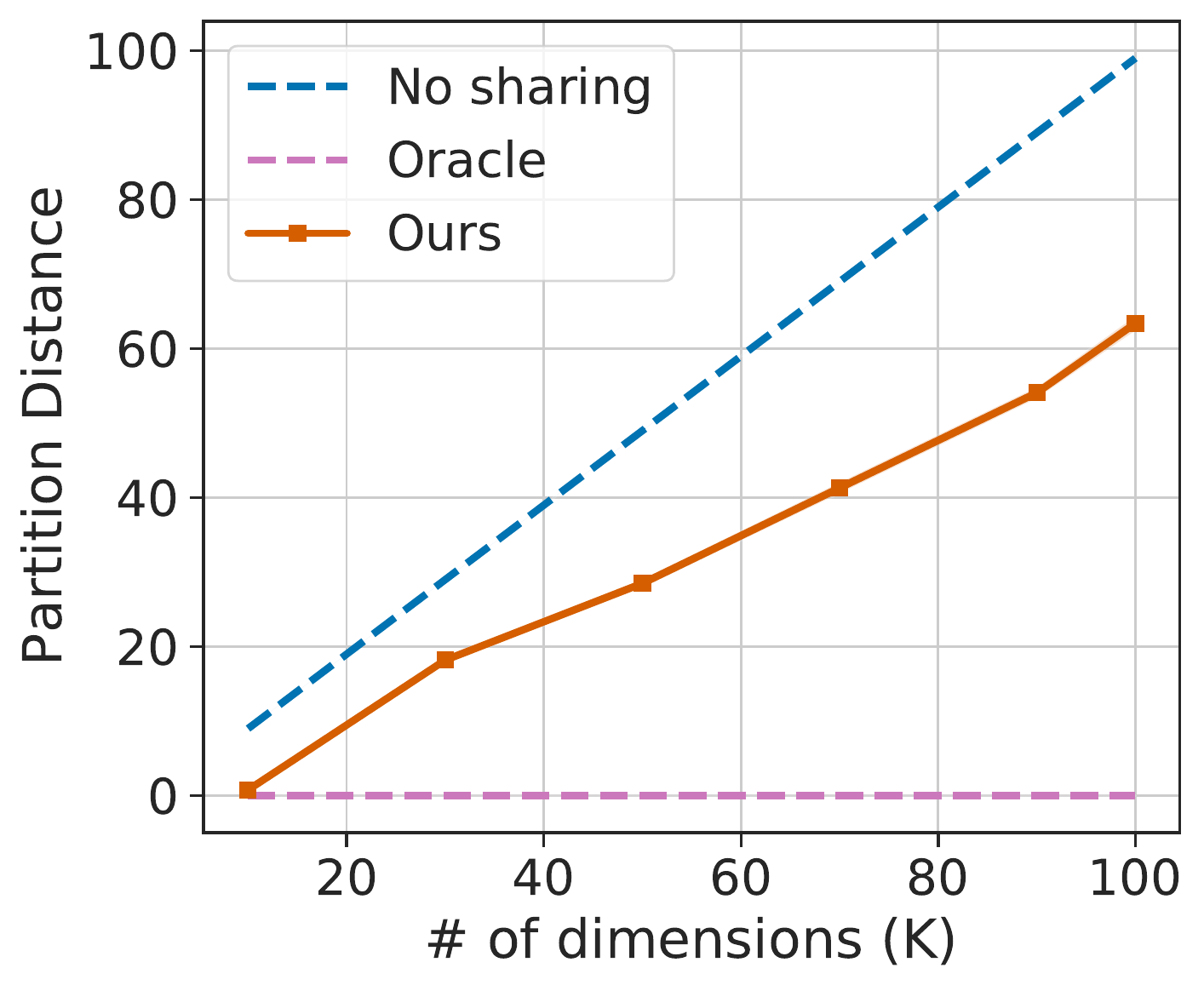}
\vspace{-0.3cm}
\caption{MSE/PD \vs \# of dimensions.} 
\label{fig:quan_ra4}
\end{minipage}
\end{figure*}
\section{EXPERIMENTS}\label{sec:exp}
We first conduct experiments on Gaussian data with random sharing as analyzed in~\secref{sec:analyze}. Next, we study recovery of known equivariances including permutation invariance and shift equivariance. Additional experimental details and results are in the Appendix.

\subsection{Gaussian Data with Shared Means}\label{sec:exp_gaussian}%
\vspace{-0.3cm}
{\noindent\bf Task,  Data and Metrics.}
We study  the same task of mean estimation that we analyzed in~\secref{sec:analyze}. 
We generate datasets of Gaussian vectors, following~\equref{eq:data_gauss}. %
We consider two evaluation metrics:
1) MSE which quantifies the performance of the mean estimation; 
2) The {\it Partition Distance (PD)}, discussed in~\secref{sec:quan_eval}, which quantifies the accuracy of the recovered sharing-scheme compared to $\mA_{\tt gt}$.

{\noindent\bf Baselines.}
We consider {\bcm No sharing} and {\bcm Oracle}. We fix $\mA$ to be the identity matrix for {\bcm No sharing}.  For {\bcm Oracle}, we directly use $\mA_{\tt gt}$. Both of these methods are trained on the entire dataset $\cD$. We also compare to {\bcm Ours*} which uses brute-force to exactly solve the bi-level optimization in~\equref{eq:struct_main}.

{\noindent\bf Results.} We conduct empirical studies over the dimension $K$, the ratio $r$ between sizes of training and overall dataset, the amount of sharing $\Rank{\mA_{\tt gt}}$, and lastly scalability of our approach when $K$ is large. 
We report the mean and shade the 95\% confidence interval computed from 200 runs for all experiments. 

First, we empirically study the method's performance over the number of dimensions with data generated for $\Rank{\mA_{\tt gt}}=1$. %
In~\figref{fig:quan_ra1}, we report both the MSE and Partition distance for each of the baselines. We observe that the approach consistently outperforms {\bcm No sharing} across different number of dimensions in both evaluation metrics. When comparing {\bcm Ours*} to {\bcm Ours}, solving the optimization via brute-force achieves the best result, while the relaxed optimization, {\bcm Ours}, remains competitive. 
 
Second, we evaluate the effect of  adjusting the train/val split in~\figref{fig:quan_ra2}. In this experiment $\Rank{\mA_{\tt gt}}=1$. %
We observe that the performance decreases as the percentage of training examples increases. This is consistent with the upper bound in~\equref{eq:bound} and the intuition that more validation data should be used when there is more sharing. 

Third, we study the effect of  $\Rank{\mA_{\tt gt}}$. Results are shown in~\figref{fig:quan_ra3}. In this case, the data consists of 5 dimensions with varying $\Rank{\mA_{\tt gt}}$ from 1 to 5. We observe that {\bcm Ours*} and {\bcm Ours} outperform {\bcm No sharing} in both metrics across the ranks.

Lastly, we demonstrate that the approach scales to higher dimensions. In~\figref{fig:quan_ra4}, we increase the dimensions from 10 to 100 with $\Rank{\mA_{\tt gt}}=1$. In these cases, brute-force optimization, \ie, {\bcm Ours*}, is no longer possible. Still, the relaxed optimization {\bcm Ours} consistently outperforms the {\bcm No sharing} baseline across  dimensions. 

\begin{table}[t]
\centering
\caption{Ablation study on the proposed penalty terms on Gaussian data.}
\setlength{\tabcolsep}{4.5pt}
\begin{tabular}{ccccc}
\specialrule{.15em}{.05em}{.05em}
\# dim & Entropy & Rank & MSE & PD\\
\hline
\hline
2 &\xmark &\xmark & $0.019 \pm .003$  & $.915 \pm 0.03$\\
2 &\cmark &\xmark & $0.016 \pm .003$  & $.385 \pm 0.07$ \\
2 &\cmark &\cmark & $0.014 \pm .003$  & $.145 \pm 0.05$ \\
\hline
4 &\xmark &\xmark & $0.035 \pm .003$  & $2.05 \pm .069$   \\
4 &\cmark &\xmark & $0.033 \pm .003$  & $1.68 \pm .070$   \\
4 &\cmark &\cmark & $0.025 \pm .003$  & $0.49 \pm .095$   \\
\hline
6 &\xmark &\xmark & $0.050 \pm .004 $  & $3.35 \pm .075$   \\
6 &\cmark &\xmark & $0.048 \pm .004 $  & $3.36 \pm .094$   \\
6 &\cmark &\cmark & $0.028 \pm .004 $  & $0.59 \pm .132$   \\
\specialrule{.15em}{.05em}{.05em}
\end{tabular}
\label{tab:sub_ablation}
\end{table}

{\bf\noindent Ablation Study on Proposed Penalty Functions.}
We perform an ablation study on the two regularization/penalty terms using Gaussian data with, the same setup reported in~\figref{fig:quan_ra1}. The results are summarized  in~\tabref{tab:sub_ablation}: both of the introduced penalty terms improve the performance in both MSE and PD.

\begin{figure}[t]
    \centering
    \includegraphics[height=0.19\textwidth]{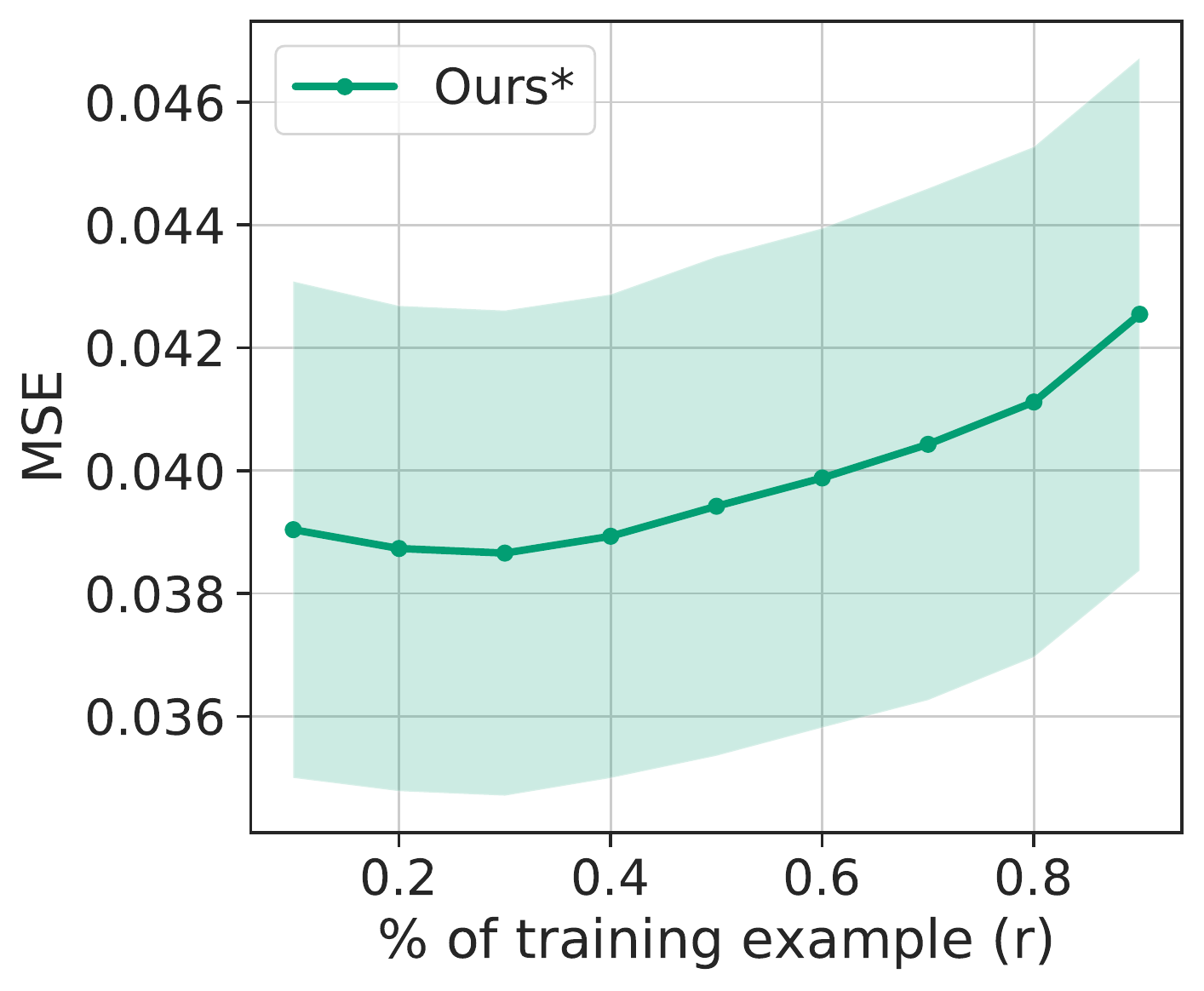}
    \includegraphics[height=0.19\textwidth]{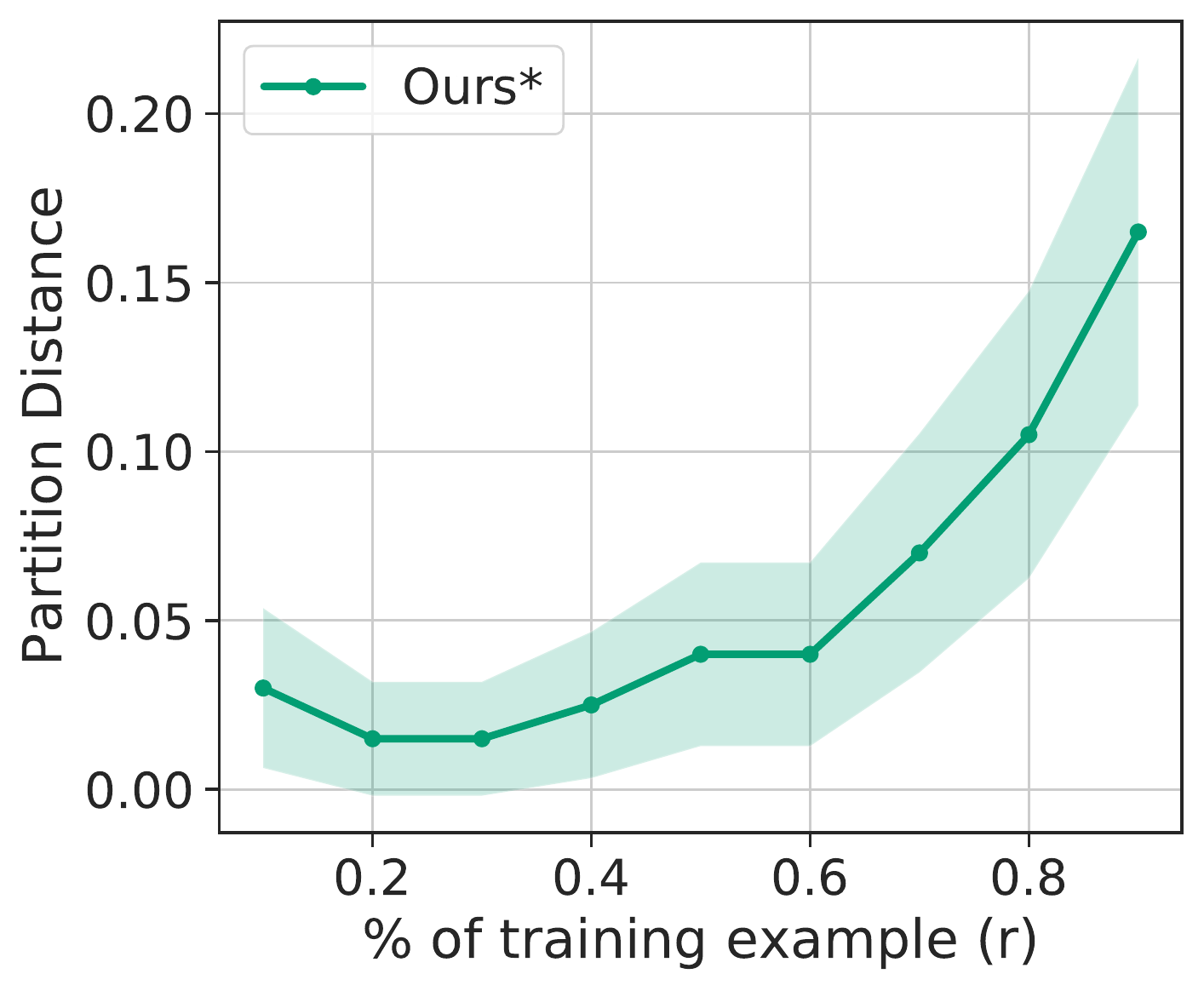}
    \vspace{-0.1cm}
    \caption{MSE/PD \vs \% of training data on data with $\Rank{\mA_{\tt gt}}=4$.}
    \label{fig:gauss_abla}
\end{figure}

{\bf\noindent Validating Theory on  Gaussian Data.}
Here, we generate data with $\Rank{\mA_{\tt gt}}=4$ to demonstrate the trade-off between the size of training and validation sets. From~\figref{fig:gauss_abla}, we can empirically observe the trade-off characterized by the theoretical upper-bound. We also observe the same trade-off in terms of partition distance (PD), which further suggests that PD is a suitable evaluation metric.

\vspace{-0.3cm}
\subsection{Recovering Permutation Invariance}
{\bf Tasks, Data and Metrics.} 
Following ~\citet{zaheer2017deep}, the task is to regress to the sum of a sequence of numbers provided in text format. \Eg, given the input (``one,'' ``five'')  the model should output 6. We also consider a variant of this task, where the ``even position'' (zero indexing) numbers are negated, \ie, an input of (``one,'' ``five'') results in $4$. Note that this latter task is only permutation invariant within the even/odd positions. The numbers are uniformly sampled from  the interval $[1,10]$ and the labels contain additive noise uniformly sampled from $[-0.5,0.5]$. 

For evaluation, we report the average squared difference ($\ell_2$-loss) between prediction and ground-truth for each of the models. As in~\secref{sec:exp_gaussian}, we also report the \textit{partition distance} to evaluate the quality of the recovered $\mA_{\tt val}$. 

{\bf Baselines.}
We consider baselines of {\bcm No sharing}, {\bcm Oracle}, and {\bcm Augerino}~\cite{benton2020learning}. For {\bcm No sharing}, the model parameters are independent. For {\bcm Oracle}, in the standard sum of numbers,
the model parameters are shared across all number positions. In the variant which negates some numbers, the model parameters are shared only across even/odd positions respectively.  For {\bcm Augerino}, we parameterize the permutation transformations using the Gumbel-Sinkhorn method~\cite{mena2018learning} to enable the training of augmentation parameters. We sample three transformations during both train and test.

\begin{figure*}[t]\centering
\setlength{\tabcolsep}{1pt}
\begin{tabular}{cc}
\textbf{Standard Sum of Numbers} & \textbf{Variant Sum of Numbers}\\
\includegraphics[height=0.19\textwidth, width=0.245\textwidth]{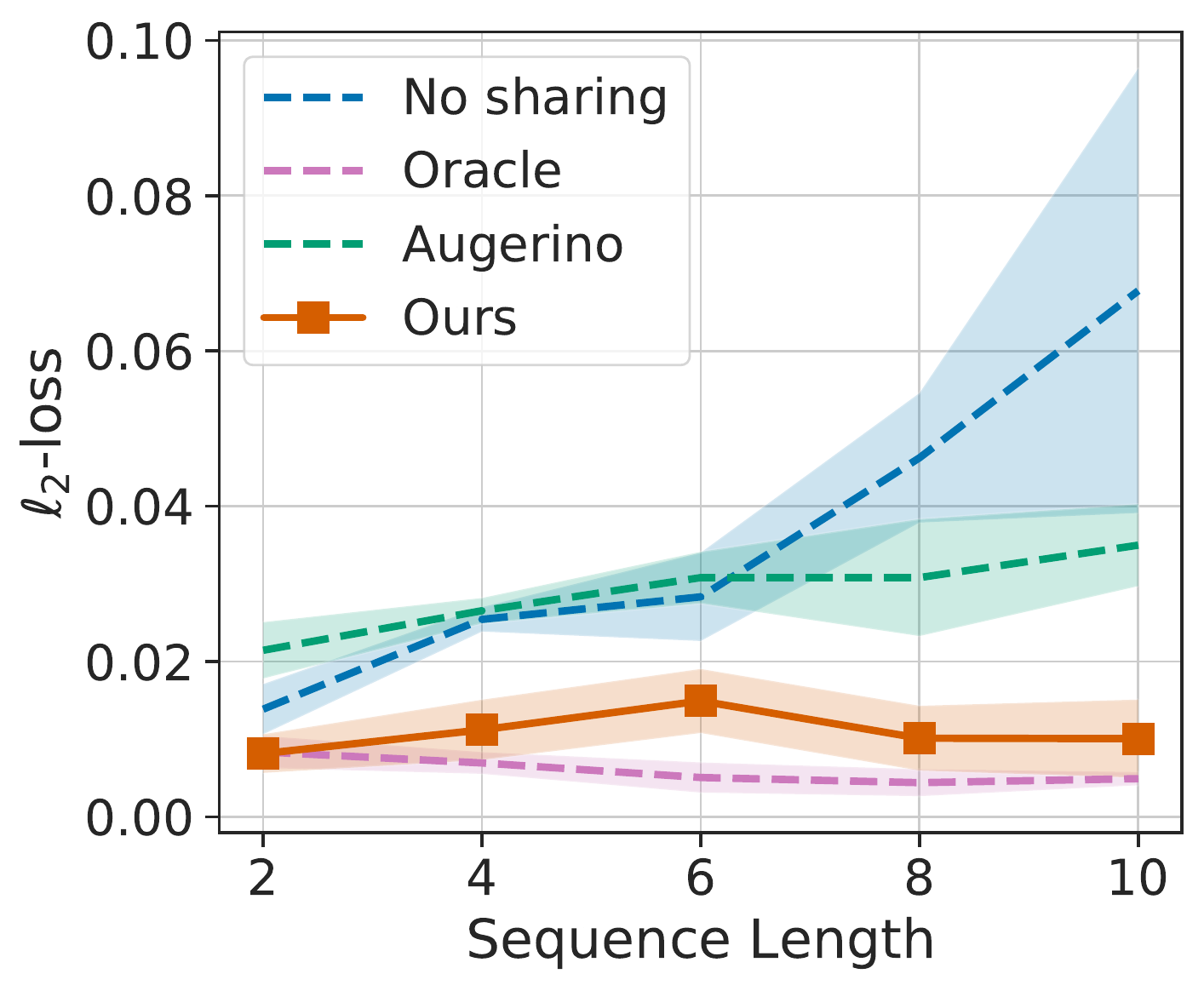}
\includegraphics[height=0.19\textwidth, width=0.245\textwidth]{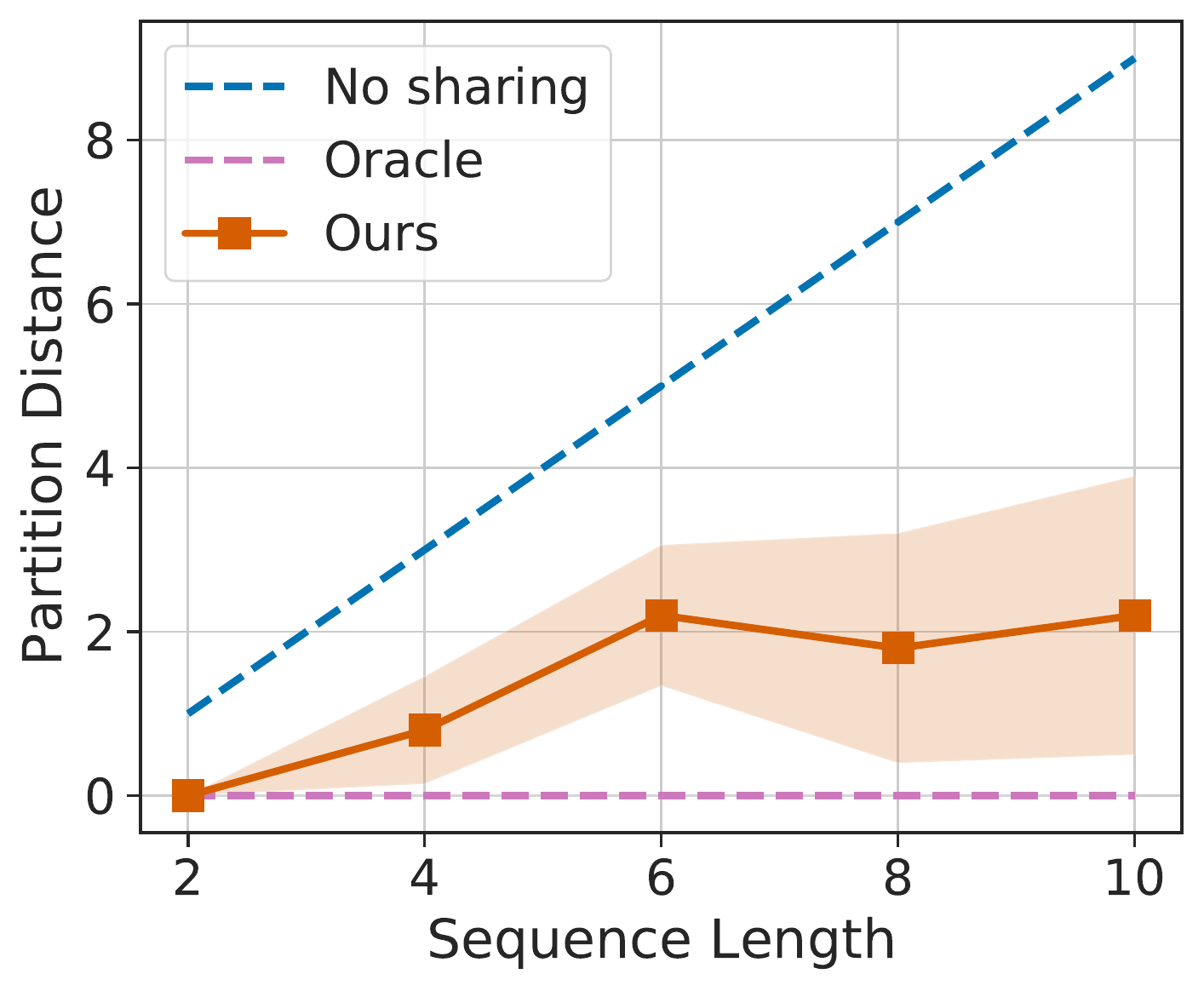} &
\includegraphics[height=0.19\textwidth, width=0.245\textwidth]{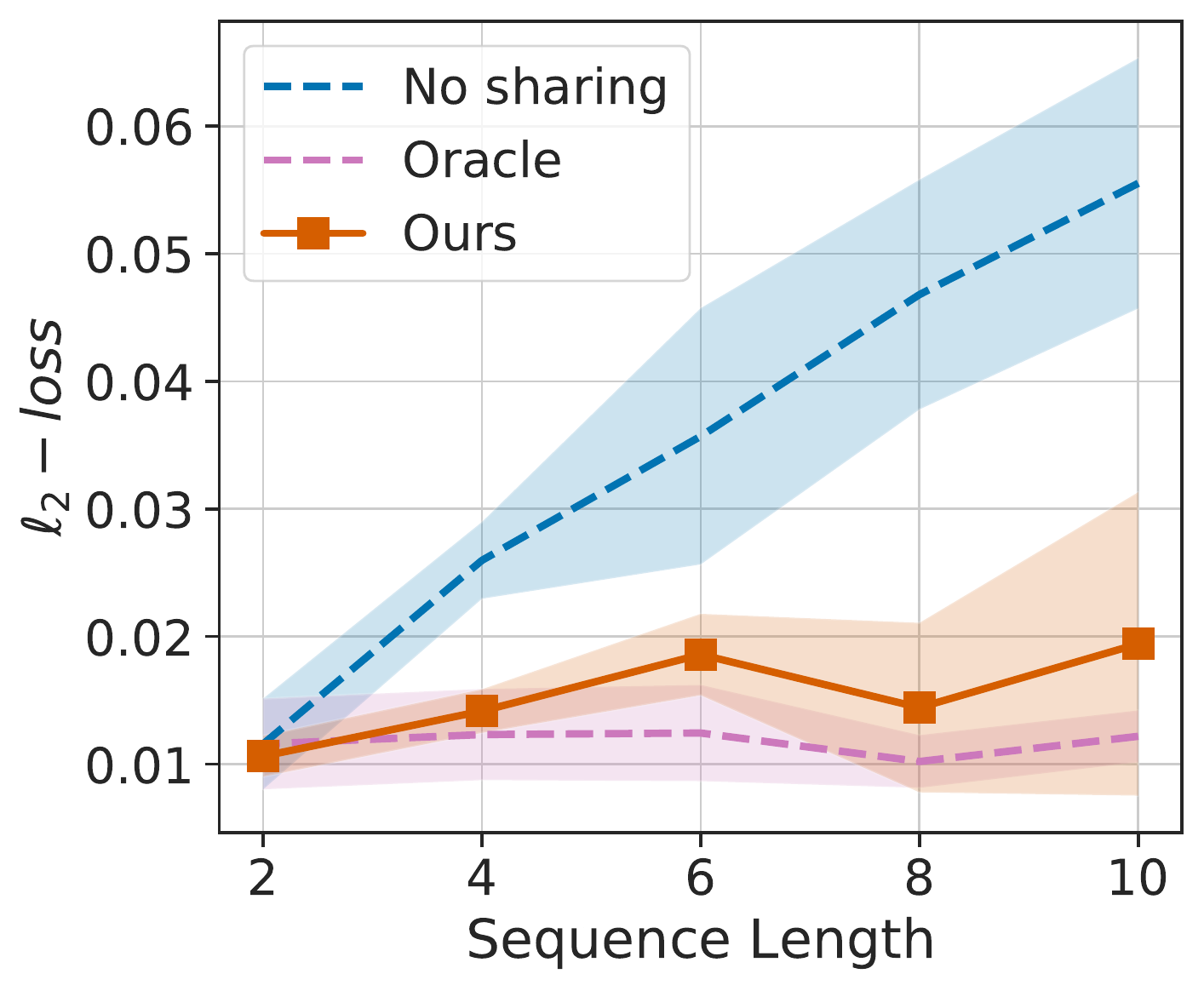}
\includegraphics[height=0.19\textwidth, width=0.245\textwidth]{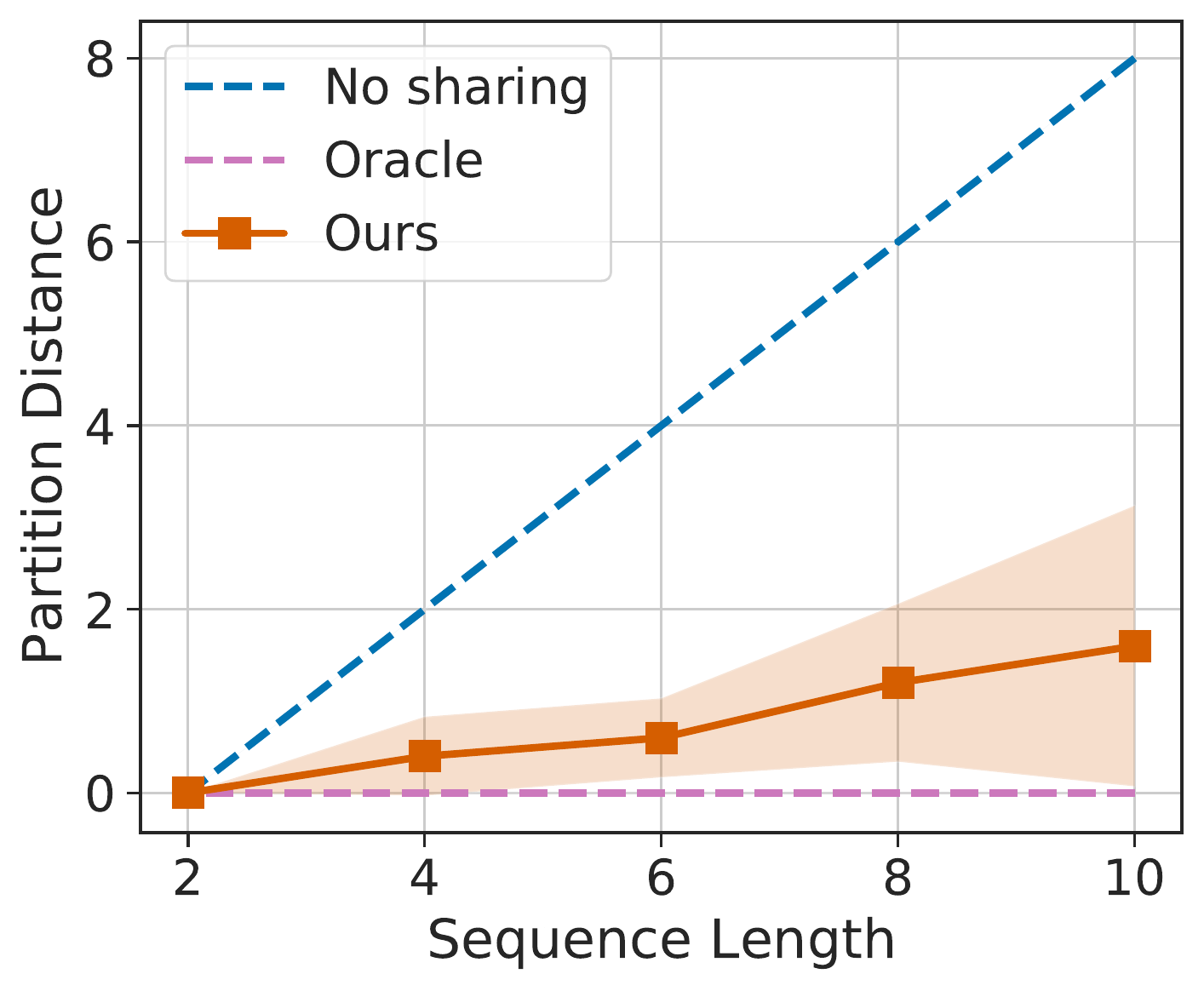}
\end{tabular}
\vspace{-0.3cm}
\caption{$\ell_2$-loss and partition distance results on standard and variant sum of numbers.}
\label{fig:quan_pa2}
\end{figure*}

{\bf Results.}
We report quantitative results across sequence length for standard sum of numbers and negated sum of numbers in~\figref{fig:quan_pa2}. All results are averaged over 5 runs using different random seeds to generate the data. We report the mean and $95\%$ confidence interval. 

In the standard sum of numbers, {\bcm Ours} outperforms the {\bcm No sharing} baseline and {\bcm Augerino}, and performs on par with {\bcm Oracle} in terms of  $\ell_2$-loss. Next, looking at the partition distance, {\bcm Ours} successfully recovers permutation invariance for sequence length of two and four. While the  partition distance increases for longer sequences, the model did learn partial permutation invariance among the number positions. 

Next, for the negated sum of number variant, {\bcm Augerino} did not learn a competitive model. The main challenge: the probability of sampling a permutation matrix that leads to the correct invariance is low. The model seems sensitive to Gumbel Sinkhorn's temperature term. In contrast, {\bcm Ours} outperforms  {\bcm No sharing} and is competitive to {\bcm Oracle}.

\begin{figure}
\centering
\vspace{-0.1cm}
\includegraphics[height=0.18\textwidth, width=0.24\textwidth]{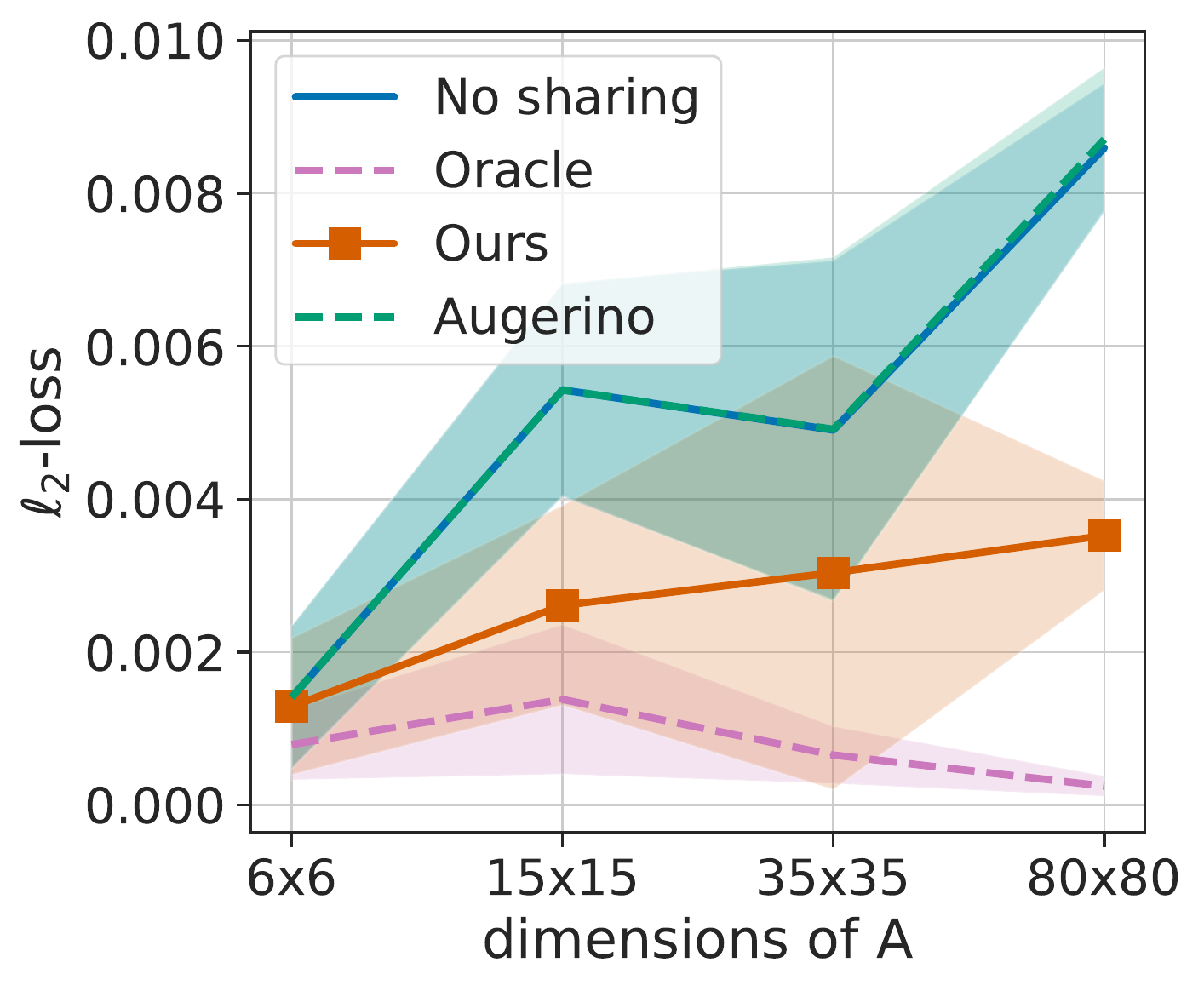}
\hspace{-0.265cm}
\includegraphics[height=0.18\textwidth, width=0.24\textwidth]{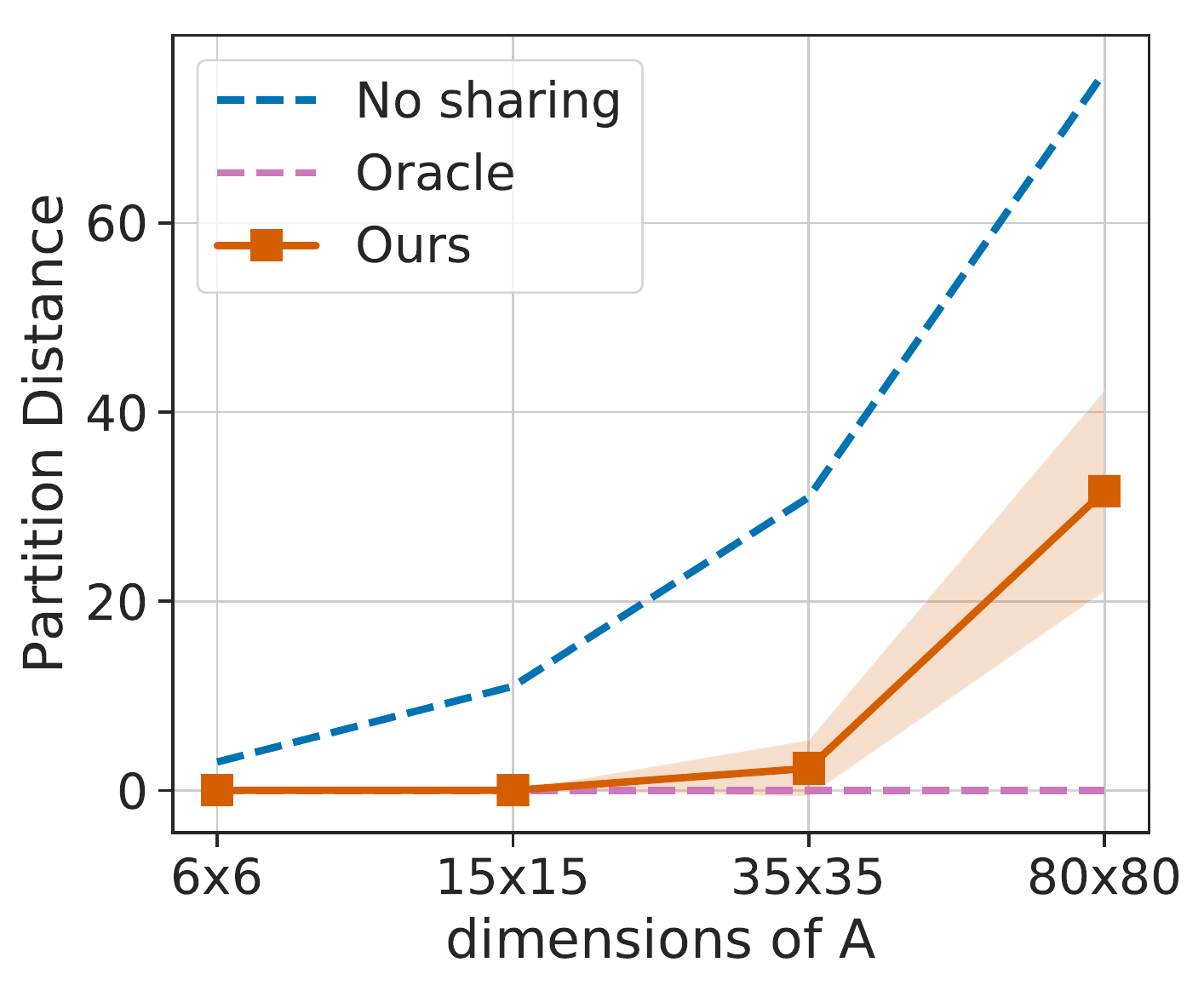}
\vspace{-.5cm}
\caption{Results for cross-correlation.}
\label{fig:quan_shift}
\end{figure}

\subsection{Recovering Shift Equivariance}
\textbf{Task, Data and Metrics.}
The task is to regress to the output of the cross-correlation operation with additive Gaussian noise, \ie, %
\bea
\rvy[k] = \eps+ \sum_{j=0}^{G-1} \rvx[k+j]\rvg[j],
\eea
where $\eps\sim \cN(0,0.1)$ and $\rvg \in \mathbb{R}^{G}$ denotes a 1D kernel.
We sample the input $\rvx \in \mathbb{R}^{K}$ from a Gaussian distribution; noise is not added for the test set. 
Note that cross-correlation is equivariant to shifts and is a linear system, $\rvy = \mG\rvx$, where $\mG$ is a Toeplitz matrix. 

We investigate  whether the studied approach  recovers this sharing scheme, \ie, $\mA_{\tt val}\bm\psi = \text{Flatten}(\mG)$, where $G$ is a Toeplitz matrix. We report the {\it $\ell_2$-loss} between the prediction and the label. We also report {\it partition distance} following~\secref{sec:exp_gaussian}.

\textbf{Baselines.} We consider baselines: {\bcm No sharing}, {\bcm Oracle}, and {\bcm Augerino}~\cite{benton2020learning}. 
For {\bcm Augerino}, we use their augmentation over the set of shift transformations with five augmented samples.

\textbf{Results.}
In~\figref{fig:quan_shift}, we report the {\it $\ell_2$-loss} and partition distance for each of the models. We observe {\bcm Ours} to outperform baselines {\bcm No sharing} and {\bcm Augerino} in terms of {\it $\ell_2$-loss}. Due to padding at the boundaries, {\bcm Augerino} learns not to shift the data, hence the performance is similar to {\bcm No sharing}. Next,  we observe that {\bcm Ours} can fully recover the sharing scheme for shift equivariance when the dimension of $\mA$ is $6\times6$ and $15 \times15$, achieving a partition distance of 0. For a larger $\mA$,~\eg, $35\times35$ and $80\times 80$, recovering the sharing scheme is much more challenging. In this case, {\bcm Ours} can partially recover the sharing scheme.

\section{CONCLUSION}\label{sec:conc}
We cast the process of equivariance discovery as an optimization over the parameter-sharing schemes. We analyze the proposed method using Gaussian data and provide a bound on the MSE gap. We illustrate that the approach can lead to better generalization than standard maximum likelihood training.  We discuss practical considerations useful for solving the proposed optimization. 
We also propose to use the partition distance (PD) to quantitatively evaluate the recovered sharing schemes and show how PD is related to the symmetric difference between two equivariant groups. Through experiments, we demonstrate that the approach can recover known equivariance properties and PD is a useful evaluation metric.

{\noindent \bf Limitations.}
Our theoretical analysis assumes Gaussian distributions which may not be met in practice. Also, the approach considers discrete group actions. However, we think the formulation and study pave the way to future research in equivariance discovery.

\subsubsection*{Acknowledgments}
We thank NVIDIA for providing GPUs used for this work. This work was supported in part by NSF under Grant \#1718221, 2008387, 2045586, 2106825, MRI \#1725729, NIFA award 2020-67021-32799 and Cisco Systems Inc.\ (Gift Award CG 1377144 - thanks for access to Arcetri). RY is supported by a Google PhD Fellowship.

\bibliography{param_share}
\bibliographystyle{icml2021}

\clearpage
\appendix
\onecolumn \makesupplementtitle
\appendix
\renewcommand{\thetable}{A\arabic{table}}
\setcounter{table}{0}
\setcounter{figure}{0}
\renewcommand{\thetable}{A\arabic{table}}
\renewcommand\thefigure{A\arabic{figure}}

This appendix is organized as follows:
\begin{itemize}
\item In~\secref{sec:supp_vb_proof}, we provide the full proof of~\clmref{clm:vb}.
\item In~\secref{sec:supp_gauss_proof}, we provide the full proof of~\clmref{clm:bound}.
\item In~\secref{sec:supp_pd_proof}, we provide the full proof of~\clmref{clm:pd}.
\item In~\secref{sec:supp_results}, we provide an ablation study and additional experimental results.
\item In~\secref{sec:supp_background}, we provide additional background and proof details.
\item In~\secref{sec:supp_exp}, we discuss experimental and implementation details for our empirical results.
\item In~\secref{sec:supp_code}, we provide link to our code.
\end{itemize}

\section{Proof of~\clmref{clm:vb}}
\label{sec:supp_vb_proof}
\vspace{6pt}
\begin{mdframed}[style=MyFrame]
\vspace{0.1cm}
\vb*
\vspace{0.1cm}
\end{mdframed}

\begin{proof} 
Let $S_i$ denote the set of indices that share parameters for the $i^{\text{th}}$ dimension of $\hat{\bm\theta}_{\tt val}$ as characterized in $\mA_{\tt val}$. 
Formally, $S_i \triangleq \{k \in \{1,\hdots, K\}|\; \forall j\; \mA_{\tt val}[i,j]=1 \land \mA_{\tt val}[k,j]=1\}$. 
As the dimensions and samples are independent, the maximum likelihood estimator is the average over the shared dimensions and samples, \ie,
\be 
\tag{\ref{eq:bias_sub}}
\mathbb{E}(\hat{\bm\theta}_{\tt val}[i]) = \frac{1}{|S_i|}\sum_{k \in S_i} {\bm \theta_{\tt gt}[k]},
\ee
Similarly, the variance is\vspace{-0.3cm}
\be 
\tag{\ref{eq:var_sub}}
\mathbb{V}(\hat{\bm \theta}_{\tt val}[i]) = \frac{\sigma^2}{|S_i||\gD|}.
\ee
Now substitute these into the MSE definition 
$$\text{MSE}(\hat{\bm\theta}(\gD)) \triangleq \mathbb{E}\norm{\hat{\bm\theta}(\gD)-\bm\theta_{\tt gt}}^2
 = \norm{\text{Bias}(\hat{\bm\theta}(\gD))}^2 + \text{Trace}(\mathbb{V}(\hat{\bm\theta}(\gD))).$$

For the bias term, we can verify that $\mA\bar{\mA}_{\tt val}^{\intercal}{\bm\theta}_{\tt gt}[k] = \frac{1}{|S_i|}\sum_{k \in S_i} {\bm \theta_{\tt gt}[k]}.$
For the variance term, each independent parameter has a variance of $\frac{\sigma^2}{|S_i||\gD|}$, in total we have $\Rank{\mA_{\tt val}}$ independent parameters, hence, $\frac{\Rank{\mA_{\tt val}}\sigma^2}{|\gD|}$.
\end{proof}
\clearpage
\section{Proof of~\clmref{clm:bound}}
\label{sec:supp_gauss_proof}
\vspace{6pt}
\begin{mdframed}[style=MyFrame]
\bound*
\end{mdframed}

\begin{proof}
We first decompose~\equref{eq:mse_gap} into three parts:
\bea
\label{eq:mse_gap1}
 & &\MSE{\hat{\bm \theta}_{\tt val}(\gD)} -  \MSE{\hat{\bm \theta}_{\tt val}(\gT)},\\
\label{eq:mse_gap2}
+& &\MSE{\hat{\bm \theta}_{\tt val}(\gT)} -  \MSE{\hat{\bm \theta}_{\tt gt}(\gT)},\\
\label{eq:mse_gap3}
+&  &\MSE{\hat{\bm \theta}_{\tt gt}(\gT)} -  \MSE{\hat{\bm\theta}_{\tt gt} (\gD)}.
\eea
We then prove the claim by upper bounding ~\equref{eq:mse_gap1},~\equref{eq:mse_gap2}, and~\equref{eq:mse_gap3}. 

\textbf{Bounding \equref{eq:mse_gap1}.} Substituting results from~\clmref{clm:vb},
\bea
 &&\hspace{-1.5cm} \MSE{\hat{\bm \theta}_{\tt val}(\gD)} - \MSE{\hat{\bm \theta}_{\tt val}(\gT)}\\
&&\hspace{-0.5cm}= \frac{\Rank{\mA_{\tt val}}\sigma^2}{|\gD|} -  \frac{\Rank{\mA_{\tt val}}\sigma^2}{|\gT|}\\
&&\hspace{-0.5cm}= - \frac{1-r}{r|\gD|} \Rank{\mA_{\tt val}} \sigma^2 \leq - \frac{1-r}{r|\gD|}\sigma^2.
\eea

\textbf{Bounding \equref{eq:mse_gap2}.}
We further decompose~\equref{eq:mse_gap2} into three parts via
\bea \nonumber
&&\hspace{-1.5cm}\MSE{\hat{\bm \theta}_{\tt val}(\gT)} - \MSE{\hat{\bm \theta}_{\tt gt}(\gT)}\\
\label{eq:mse_gap2a}
&&\hspace{-0.5cm}=\MSE{\hat{\bm \theta}_{\tt val}(\gT)} - \hMSE{\hat{\bm \theta}_{\tt val}(\gT)} \\
\label{eq:mse_gap2b}
&&+\hMSE{\hat{\bm \theta}_{\tt val}(\gT)} - \hMSE{\hat{\bm \theta}_{\tt gt}(\gT)} \\
\label{eq:mse_gap2c}
&&+\hMSE{\hat{\bm \theta}_{\tt gt}(\gT)} - \MSE{\hat{\bm \theta}_{\tt gt}(\gT)},
\eea
and bound each term.

For \equref{eq:mse_gap2a}:
By reverse triangle inequality
\bea
&\MSE{\hat{\bm \theta}_{\tt val}(\gT)} - \hMSE{\hat{\bm \theta}_{\tt val}(\gT)}\\
=& \norm{\mathbb{E} \hat{\bm \theta}_{\tt val}(\gT)-{\bm \theta_{\tt gt}}}^2 - \norm{\mathbb{E} \hat{\bm \theta}_{\tt val}(\gT)-\hat{\bm \theta}_{\tt ind}(\gV)}^2\\
\leq& \norm{-{\bm \theta}_{\tt gt} + \hat{\bm \theta}_{\tt ind}(\gV)}^2 = 
\norm{{\bm \theta}_{\tt gt} - \hat{\bm \theta}_{\tt ind}(\gV)}^2.
\eea

Let $\mZ=\bm \theta_{\tt gt} - \hat{\bm\theta}_{\tt ind}(\gV)$, then $\mZ \sim \cN({\bm 0}, \frac{\sigma^2}{|\gV|}\mI)$. This means 
\bea
U = \frac{\norm{\mZ}^2}{\frac{\sigma^2}{\cV}}=\frac{\sum_{i}^K Z_i^2}{\frac{\sigma^2}{|\gV|}} \sim \chi^2_k.
\eea

From the tail bound of the $\chi^2$ distribution~\cite{laurent2000adaptive} and $t\geq1$,
\bea
P(U \geq 2tK) \leq \exp\left(-\frac{tK}{10}\right).
\eea
Therefore, with probability $1-\alpha$ where $\alpha\leq\exp(-\frac{tK}{10})$,
\bea
&\hMSE{\hat{\bm \theta}_{\tt val}(\gT)} - \MSE{\hat{\bm \theta}_{\tt val}(\gT)}\\ 
\leq& -20\ln(\alpha)\frac{\sigma^2}{|\gV|} = -20\ln(\alpha)\frac{\sigma^2}{(1-r)|\gD|}.
\eea

For \equref{eq:mse_gap2b}: As $\mA_{\tt val}$ is determined on the validation set, it has the smallest $\widehat{\text{MSE}}$, hence 
\bea
\hMSE{\hat{\bm \theta}_{\tt val}(\gT)} - \hMSE{\hat{\bm \theta}_{\tt gt}(\gT)} \leq 0. 
\eea

For \equref{eq:mse_gap2c}: Similar to~\equref{eq:mse_gap2a} we obtain
\bea
&\hMSE{\hat{\bm \theta}_{\tt gt}(\gT)} - \MSE{\hat{\bm \theta}_{\tt gt}(\gT)}\\
=& \norm{\mathbb{E} \hat{\bm \theta}_{\tt gt}(\gT)-\hat{\bm \theta}_{\tt ind}(\gV)}^2
= \norm{\bm \theta_{\tt gt} - \hat{\bm\theta}_{\tt ind}(\gV)}^2.
\eea

\textbf{Bounding \equref{eq:mse_gap3}.} Substituting results from~\clmref{clm:vb},
\bea
&\MSE{\hat{\bm \theta}_{\tt gt}(\gT)} - \MSE{\hat{\bm\theta}_{\tt gt} (\gD)}\\
=& \frac{\Rank{\mA_{\tt gt}} \sigma^2}{|\gT|} - \frac{\Rank{\mA_{\tt gt}} \sigma^2}{|\gD|}\\
=& \frac{1-r}{r|\gD|} \Rank{\mA_{\tt gt}} \sigma^2.
\eea
Summing up the individual bounds concludes the proof.
\end{proof}

\section{Proof of~\clmref{clm:pd}} \label{sec:supp_pd_proof}
\vspace{6pt}
\begin{mdframed}[style=MyFrame]
\pd*
\end{mdframed}
\begin{proof} 
Let  $\gP_K^{(1)}$ and $\gP_K^{(2)}$ denote the corresponding partitions of $\mA^{(1)}$ and $\mA^{(2)}$. Let $j$  be an element that must be moved in $\gP_K^{(1)}$ to match $\gP_K^{(2)}$. Let $\gC_{j*}^{(1)}$ and $\gC_{j*}^{(2)}$ refer to the cluster that contains the element $j$. 

\textbf{Lower bound.} As $j$ must be moved, this means that $\gC_{j*}^{(1)}$ are $\gC_{j*}^{(2)}$ are not identical, therefore, $|\Pi_{\gC_{j*}^{(1)}} \;\Delta\; \Pi_{\gC_{j*}^{(2)}}| \geq 1$. 
As $\Pi_{\gC_{j*}^{(1)}}  \subseteq  \gG_{K,K}^{(1)} $ and $\Pi_{\gC_{j*}^{(2)}}  \subseteq  \gG_{K,K}^{(2)} $, \underline{for each $j$} there exist a difference of at least 1. Therefore, $$|\gG_{K,K}^{(1)} \Delta \gG_{K,K}^{(2)} | \geq PD(\mA^{(1)}, \mA^{(2)}).$$

\textbf{Upper bound.} Similarly,
we can upper bound $|\Pi_{\gC_{j*}^{(1)}} \;\Delta\; \Pi_{\gC_{j*}^{(2)}}|$ with $K!-1$. Consider the case, when $\gC_{j*}^{(1)}$ contains all the elements and $\gC_{j*}^{(2)}$ contains only $j$. 
As $\Pi_{\gC_{j*}^{(1)}}  \subseteq  \gG_{K,K}^{(1)} $ and $\Pi_{\gC_{j*}^{(2)}}  \subseteq  \gG_{K,K}^{(2)} $, \underline{for each $j$} there exist a difference of at most $K!-1$. Therefore, $$(K!-1)  \cdot PD(\mA^{(1)}, \mA^{(2)}) \geq \gG_{K,K}^{(1)} \Delta \gG_{K,K}.$$%
\end{proof}

\clearpage

\section{Additional Results}\label{sec:supp_results}
\subsection{Recovering Shift Equivariance from Denoising} 
{\bf Task, data and metrics.} The task is to denoise 1D signals with additive Gaussian noise, \ie, 
\vspace{-0.1cm}
\be\label{eq:denoise}
\hat{\rvy} = f_\theta(\rvx),
\vspace{-0.05cm}
\ee
where $\rvx\in\mathbb{R}^{K}$ is the noisy signal, $f_\theta$ is the denoising function and $\rvy\in\mathbb{R}^{K}$ is the clean signal. We denoise using a linear model, \ie, $\hat{\rvy} = \mG\rvx$.

We create the data by adding Gaussian noise to a randomly scaled and translated unit step signal, 
\be
\rvx[k] = \underbrace{s \cdot U(k-t) +b}_{\rvy[k]} + \epsilon,
\ee
where $U$ denotes the unit step function,  $s\sim \text{unif}[1,50]$, $b\sim \text{unif}[-5,5]$, $t\sim \text{unif}\{0, K\}$ and $\epsilon$ is  zero-mean Gaussian noise. 
We report the mean squared error (MSE) between the prediction and the clean ground truth  signal. The training set $\gT$ consists of 50 examples, the validation set $\gV$ consists of 100 examples, and we use 10,000 examples for testing.

{\bf Baselines.} We consider two baselines: {\bcm No sharing} and {\bcm Augerino}~\cite{benton2020learning}. 
For {\bcm Augerino}, we use their augmentation over the set of shift transformations with five augmented samples, \ie, during both train and test, this method requires five forward passes for each input.

\textbf{Implementation details.} In this experiment, we consider learning the linear system as described above and minimize the {\it $\ell_2$-loss}. Again, the lower-level task is solved analytically. We use the Adam optimizer for the upper-level optimization with a learning rate of 0.2. 

\begin{figure}[t]
\centering
\begin{minipage}{0.4\textwidth}
\centering
\begin{tabular}{cc}
\\
\includegraphics[height=4.5cm]{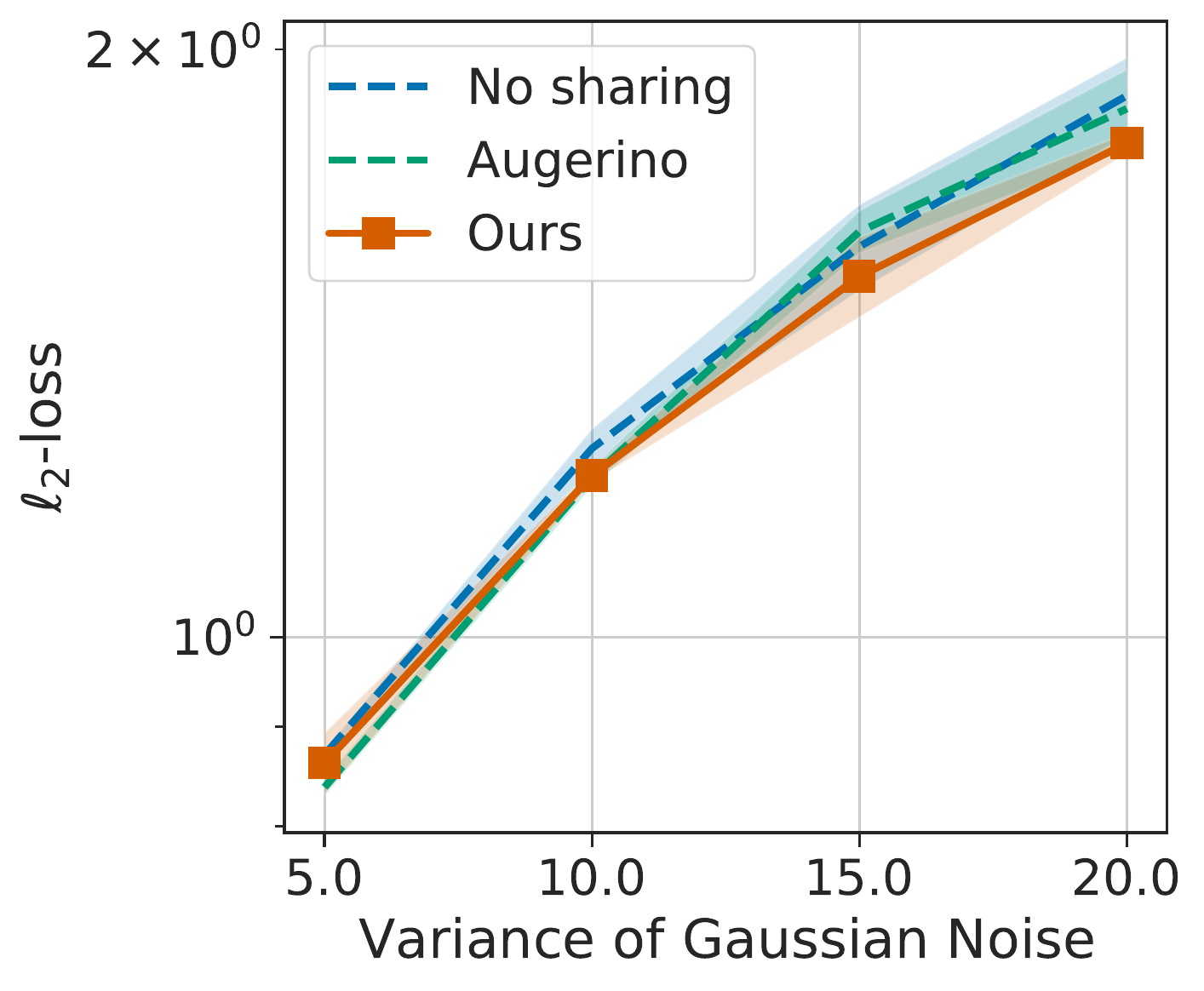}
\end{tabular}
\vspace{-0.3cm}
\caption{Quantitative results for denoising.}
\label{fig:supp_quan_denoise}
\end{minipage}
\hspace{0.3cm}
\begin{minipage}{0.55\textwidth}
\centering
\begin{tabular}{cc}
{\bcm No sharing} & {\bcm Ours}\\
\includegraphics[height=4.3cm]{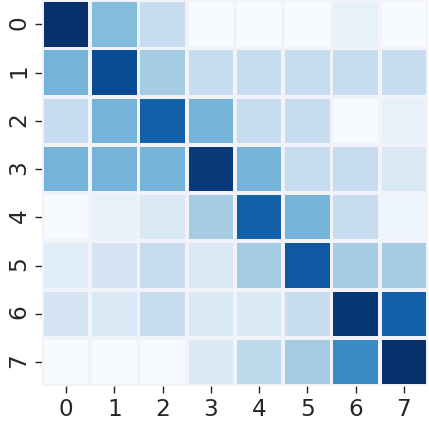} & 
\includegraphics[height=4.3cm]{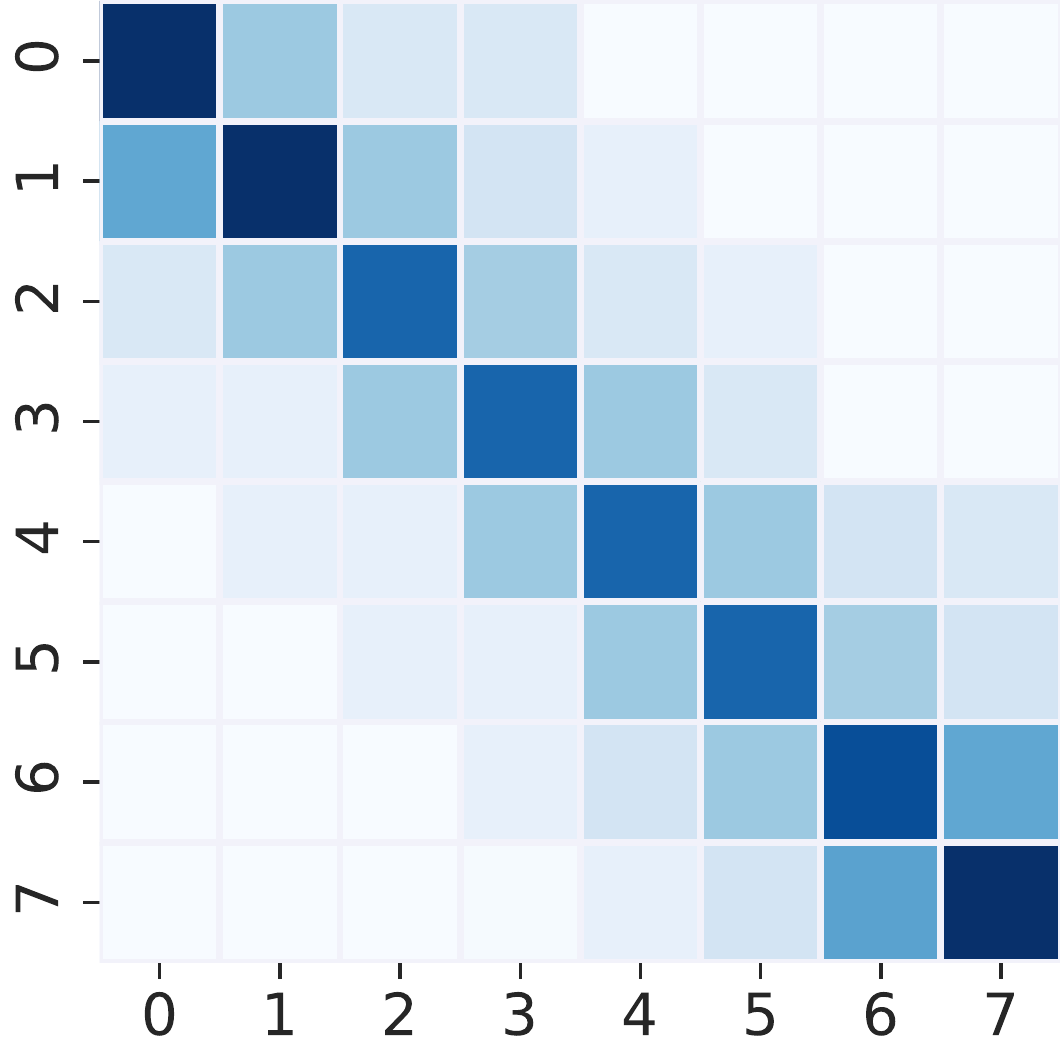}\\
\end{tabular}
\vspace{-0.3cm}
\caption{Visualization of the learned $\mG$.}
\label{fig:supp_qual_denoise}
\end{minipage}
\end{figure}

{\bf Results.}
In~\figref{fig:supp_quan_denoise}, we report the {\it $\ell_2$-loss} across different amounts of added noise. This is averaged over five runs with different random seeds and we plot the mean and 95\% confidence interval. We observe larger gains of {\bcm Ours} over the baselines when there is more noise to be removed. When there is less noise, {\bcm No sharing}, {\bcm Augerino} and {\bcm Ours} are comparable. 

In~\figref{fig:supp_qual_denoise}, we visualize the learned $\mG$ for {\bcm No sharing} and {\bcm Ours}. We observe that our approach successfully learns a Toeplitz matrix capturing the shift equivariance property of the data. This is not the case for {\bcm No sharing}.

\subsection{Additional Comparison with Augerino}
In our paper, we reported Augerino with three or five transformations. For completeness, we report additional experiments using more transformations. In~\tabref{tab:sub_timing} we  report the training/testing time for standard sum of numbers with sequence length of 10.
As can be seen, the performance of Augerino improves with increased number of transformations; similarly for the memory usage and inference time. The inference time is measured on an NVIDIA Titan X (Pascal) averaged over 100 runs.

\begin{table}[H]
\centering
\begin{tabular}{ccccc}
\specialrule{.15em}{.05em}{.05em}
Method & Number of Trans. & $\ell_2\text{-loss}$ & Inference time & Memory Usage\\
\hline
\hline
Augerino &        3         & $0.03498 \pm 0.00528$ &   4.00 ms      &    727MiB   \\
Augerino &        15        & $0.02567 \pm 0.00370$ &   5.13 ms      &   1101MiB   \\
Augerino &        75        & $0.01966 \pm 0.00503$ &   22.0 ms      &   2965MiB   \\
Ours     &        -         & $0.01005 \pm 0.00503$ &   1.73 ms      &    645MiB   \\

\specialrule{.15em}{.05em}{.05em}
\end{tabular}
\caption{Comparison with Augerino over more transformations.}
\label{tab:sub_timing}
\end{table}

\section{Additional Background}\label{sec:supp_background}
\subsection{Parameter-sharing in cross-correlation}
Recall that a cross-correlation operation is defined via
\be\label{eq:conv_sup}
\rvy[k] = \sum_{j=0}^{G-1} \rvx[k+j]\rvg[j].
\ee
We can write this as a linear system $\rvy = \mG \rvx$, where $\mG$ is a Toeplitz matrix. To build some intuition on the parameter sharing scheme, let's consider an input $\rvx \in \mathbb{R}^3$ and $\rvg = [2,1]$. In this case $\mG$ takes the following form,
\begin{figure}[H]
\centering
\begin{minipage}[c]{0.05\textwidth}
$\mG =$
\end{minipage}
\begin{minipage}[c]{0.25\textwidth}
\includegraphics[height=2.5cm]{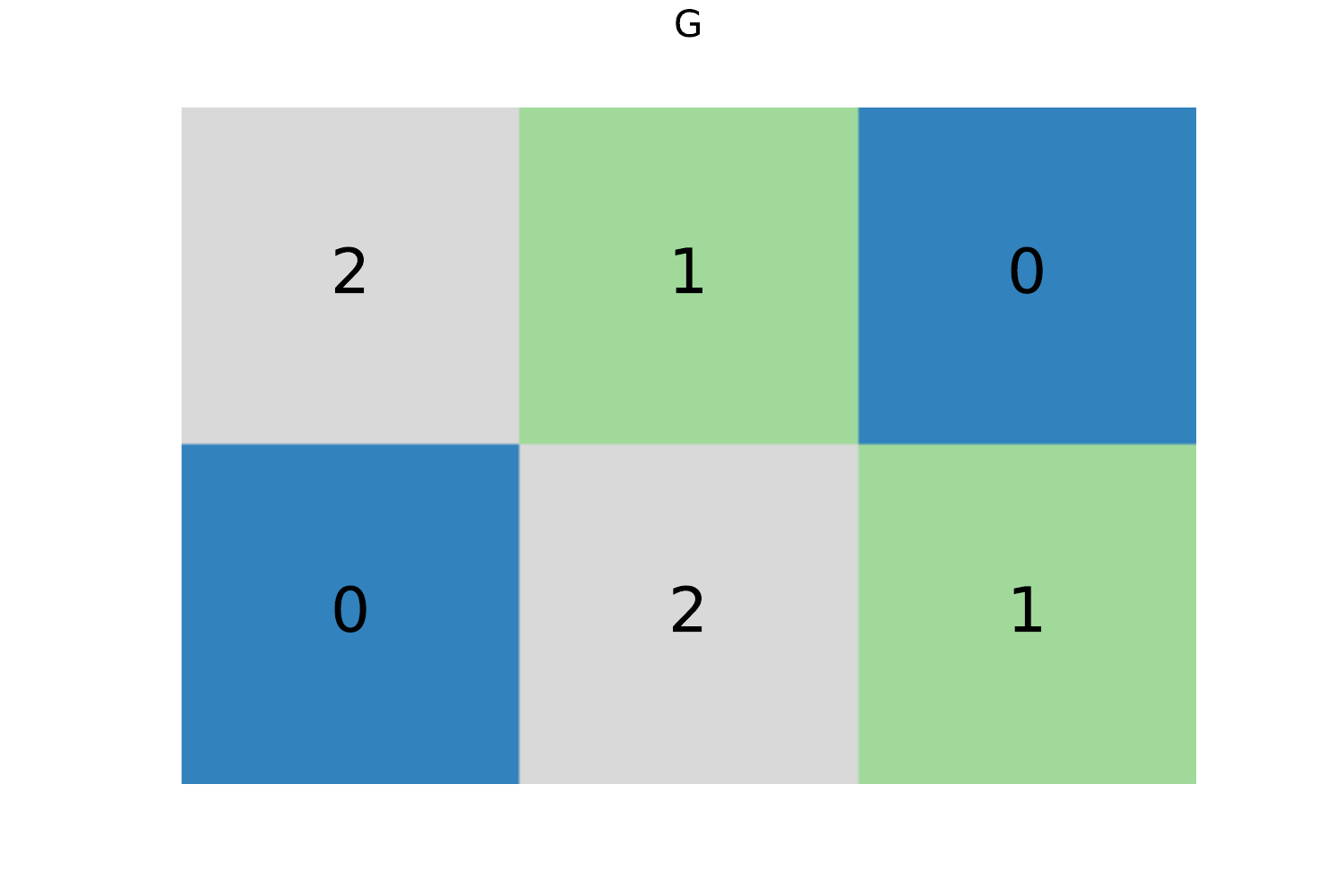}.
\end{minipage}
\end{figure}

\begin{figure}[t]
\centering
\begin{tabular}{cccc}
\includegraphics[height=3.3cm]{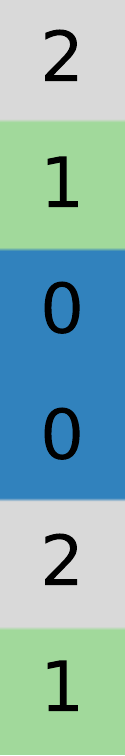} & 
& \includegraphics[height=3.3cm]{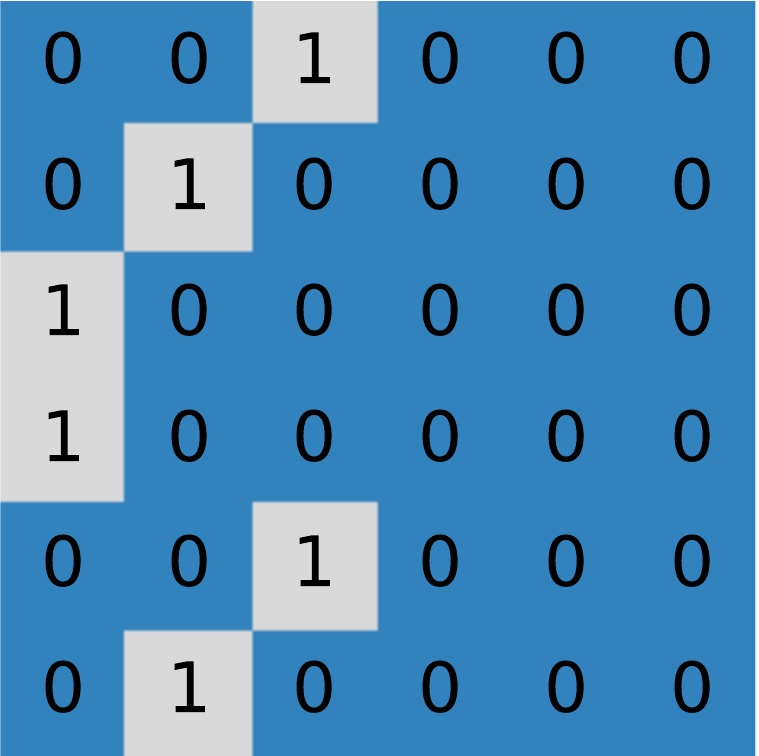} & \includegraphics[height=3cm]{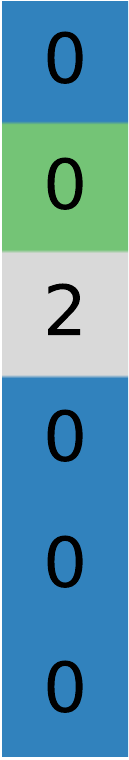}\\
Flatten($\mG$) & = & $\mA$ & $\bm\psi$
\end{tabular}
\caption{Parameter-sharing scheme for a cross-correlation.}
\label{fig:conv_drawing}
\end{figure}

Observe that the parameters are shared across rows of $\mG$. To capture this sharing scheme, we use an assignment matrix $\mA$, \ie, $\mG = \mA\bm\psi$ as illustrated in~\figref{fig:conv_drawing}, where we have flattened the matrix $\mG$ into a vector. Observe that $\mA$ selects the parameters from $\bm\psi$ to form $\mG$ characterizing a sharing scheme.

\subsection{Miscellaneous proof details}
{\bf Reverse Triangle Inequality.} 
Let $\rvx, \rvy \in \mathbb{R}^d$, then
$$
\norm{\rvx}-\norm{\rvy} \leq \norm{\rvx-\rvy}.
$$
\begin{proof} By triangle inequality,
$$\norm{\rvx} = \norm{\rvx - \rvy + \rvy}
\leq \norm{\rvx-\rvy} + \norm{\rvy}.$$
\end{proof}
{\bf Tail bound of $\chi^2$ distribution.}
Let  $U$ be a $\chi^2_K$ random variable and $t\geq1$ then
\bea
P(U \geq 2tK) \leq \exp\left(-\frac{tK}{10}\right).
\eea
\begin{proof}
From Lemma 1 of~\citet{laurent2000adaptive},
let $Y_i$ be \textit{i.i.d.}\ Gaussian variables, let $a_i$ be non-negative, and
\be
Z = \sum_{i=1}^D \rva_i(Y_i^2-1).
\ee
Then, %
\be
P(Z \geq 2\norm{\rva}_2\sqrt{x}+2\norm{\rva}_{\infty}x) \leq \exp(-x).
\ee
Next, let $\rva = [1,\hdots,1]$, then
\be
P(U \geq K + 2\sqrt{Kx} + 2x) \leq \exp(-x).
\ee
Let $x = \frac{tK}{10}$, we have
\be
P(U \geq K + 2K \cdot (\sqrt{t/10}+t/10) ) \leq \exp\left(-\frac{tK}{10}\right).
\ee
Lastly, we need to show
\bea
2tK \geq& K +  2K \cdot (\sqrt{t/10}+t/10) ),\\
2t-1 \geq& 2(\sqrt{t/10}+t/10).
\eea
Let $v=\sqrt{t/10}$, then we have
\be
0 \geq -9v^2 +v + 0.5,
\ee
which is true when $v \geq 0.3$, \ie, when $t\geq 0.9$.
\end{proof}

\section{Additional Experimental Details}\label{sec:supp_exp}
We provide additional details of the experiments reported in the main paper.
\subsection{Gaussian data with Shared Means}
{\bf Data.} We generate data following a Gaussian distribution with a shared mean as specified in~\equref{eq:data_gauss}. The dataset $\gD$ contains $100$ samples. We split $\gT$ and $\gV$ to contain $30$ and $70$ samples, except for experiment in~\figref{fig:quan_ra3} where we sweep over the different sizes of $\gT$ and $\gV$.

{\bf Implementation details.} For this task, the lower optimization in our proposed program (\equref{eq:struct_main}) can be solved analytically. To see this we write it in matrix form:
\bea
\min_{\bm \psi} \norm{\mX \bm\psi \mA^{\intercal} - \mY}_F^2,
\eea
where $\mX = \mathbf{1}_{N \times 1}$, $\bm\psi \in \mathbb{R}^{1 \times K}$, $\mA \in [0,1]^{K \times K}$ and $\mY \in \mathbb{R}^{N \times K}$.
In form of ordinary least-squares, %
\bea
\bm\theta^* = \argmin_{\bm \theta} \norm{\mX \bm\theta - \mY}_F^2,
\vspace{-0.1cm}
\eea
from which  we obtain $\bm\psi^* = \bm\theta^*(\mA^{\intercal})^{+}$, where $\mA^{+}$ denotes the pseudo-inverse. 
We use the Adam~\cite{kingma2015adam} to solve the upper-level optimization.

{\bf Training details.} As described in the paper, we solve the lower-level optimization analytically and use the Adam optimizer to handle the upper-level task. For the Adam optimizer, we use a learning rate of $2\mathrm{e}{-2}$ with weight-decay of $1\mathrm{e}{-4}$. We also tried lowering the learning rates to $1\mathrm{e}{-2}$ and $5\mathrm{e}{-3}$. We did not sweep over the weight-decay. These hyperparameters are used for all experiments and all models in this section. We train all models for 1000 epochs without batching. 

{\bf Running experiments.} We provide code to run all these experiments. Please refer to the code in folder $\tt projects/GaussianSharing/experiments/$. We use an NVIDIA TITAN X (Pascal) to run these experiments. 

\subsection{Recovering Permutation Invariance}
{\bf Data.} For the sum of numbers dataset, we uniformly sample numbers from the set $\{1,\dots, 10\}$ and the corresponding label is the sum of these numbers. Let $\rvx_i$ denote the $i^{\text{th}}$ number in the sequence, then the label is defined as 
\be
\rvy = \sum_i \rvx_i.
\ee
Next, for the variant sum of numbers, the label is defined as 
\be
\rvy = \sum_i (-1)^{(i+1)}  \rvx_i.
\ee
In this case, the even positioned numbers are multiplied by negative one.

For both of these experiments, we use a dataset $\cD$ of size 250 and split it into training set $\gT$ and validation set $\gV$  containing 100 and 150 samples respectively. As we use deep-nets for these experiments, we created a separate set $\gH$ of size 250 to apply early stopping and tune the learning rate. Note that all the compared methods have access to exactly the same data. Lastly, we use a test set of $100,000$ samples. For pre-processing, we standardize the label by subtracting the mean and by dividing by the standard deviation. At test-time, we scale the output back to the original range.

{\bf Model details.}
 We use an embedding of 500 dimensions to represent the input. Next, this embedding is passed through a fully-connected layer of 50 dimensions (one per position), and a ReLU non-linearity. The baselines use this same network architecture.
 To output a single scalar, we sum across the position dimension and pass it through a fully-connected layer of 1 dimension. We learn how to share the model-parameters across the sequence positions at the first fully connected layer.

{\bf Training details.} We solve both lower and upper-level optimization using the Adam optimizer, and compute the hypergradient using Neumann inverse approximation, with 20 iterations. The upper-level learning rate is $1\mathrm{e}{-2}$ and the lower-level learning rate is $1\mathrm{e}{-3}$. For every upper-level optimization step, we run 250 lower-level steps. For the learning rate, we studied  $1\mathrm{e}{-1}$, $1\mathrm{e}{-2}$ and $1\mathrm{e}{-3}$. For the number of lower-level steps, we have assessed $50, 150, 250$ steps.

{\bf Running experiments.} 
We provide code to run all these experiments. Please see the code in folder $\tt projects/PermutationSharing/experiments/$. We use an NVIDIA TITAN X (Pascal) to run these experiments.

\subsection{Recovering Shift Equivariance}
{\bf Data.} %
We fixed the kernel  $\rvg$ to increase by two for every position, \eg, a kernel with size three is $[1,3,5]$.  The training set $\gT$ consists of 50 examples, the validation set $\gV$ consists of 100 examples, and we use 10,000 examples for testing. %

\textbf{Implementation details.} In this experiment, we consider learning the linear system as described above and minimize the {\it $\ell_2$-loss}. As in the Gaussian experiment, the lower-level task can be solved analytically. We use the Adam optimizer to address the upper-level optimization. 

{\bf Training details.} We solve the lower-level optimization task analytically by formulating it as an ordinary least-squares problem, \ie,
\be
\mG^* = (\mX^{\intercal}\mX)^{-1}\mX^{\intercal}\mY,
\ee
where $\mX \in \mathbb{R}^{N \times K_{\tt in}}$ and $\mY \in \mathbb{R}^{N \times K_{\tt out}}$. With the lower-level optimization solved, we  write $\bm\psi^*$ as 
\be
\bm\psi^{*}(\mA) = \mA^+\text{Flatten}(\mG^*),
\ee
where $\mA^+$ denotes the pseudo-inverse and flatten reshapes a matrix into a vector. For the upper-level task, we  back-propagate through $\bm\psi^{*}$ to update $\mA$. In this experiment, we use the Adam optimizer with a learning rate of 0.1.

{\bf Running experiments.} We provide code to run all these experiments. Please see the code in folder $\tt projects/ConvSharing/experiments/$. We use an NVIDIA 
GeForce GTX 1080 to run these experiments.

\section{Code Release}\label{sec:supp_code}
Please find the released code at~\url{https://github.com/raymondyeh07/equivariance_discovery}.

\end{document}